\documentclass{article} 
\usepackage{arxiv_sty,times}

\usepackage{amsmath,amsfonts}

\def\eqref#1{equation~\ref{#1}}

\def\1{\bm{1}}

\DeclareMathAlphabet{\mathsfit}{\encodingdefault}{\sfdefault}{m}{sl}
\SetMathAlphabet{\mathsfit}{bold}{\encodingdefault}{\sfdefault}{bx}{n}

\newcommand{\R}{\mathbb{R}}

\DeclareMathOperator*{\argmin}{arg\,min}

\usepackage{enumitem}
\usepackage{multirow}
\usepackage{wrapfig}
\usepackage{colortbl}
\usepackage[normalem]{ulem}
\usepackage{microtype}
\usepackage{comment}
\usepackage[hang,flushmargin]{footmisc}

\setlist[itemize]{align=parleft,left=0pt,topsep=1mm,itemsep=0mm,parsep=1mm}

\definecolor{azure(colorwheel)}{rgb}{0.0, 0.5, 1.0}
\definecolor{nicegreen}{rgb}{0.0, 0.7, 0.1}
\definecolor{yw}{rgb}{0.01176, 0.5490, 0.5490}
\definecolor{ashblue}{rgb}{0.36, 0.54, 0.66}
\definecolor{ashgrey}{rgb}{0.7, 0.75, 0.71}
\definecolor{applegreen}{rgb}{0.55, 0.71, 0.0}
\definecolor{blue}{rgb}{0.0, 0.0, 1.0}
\definecolor{postechred}{rgb}{0.784, 0.003, 0.313}
\definecolor{ywg}{rgb}{0.9960, 0.8984, 0.5859}
\definecolor{ballblue}{rgb}{0.13, 0.67, 0.8}
\definecolor{cornellred}{rgb}{0.7, 0.11, 0.11}
\definecolor{darkcyan}{rgb}{0.0, 0.55, 0.55}
\definecolor{CuGray}{gray}{0.9}
\definecolor{airforceblue}{rgb}{0.36, 0.54, 0.66}
\definecolor{rev}{rgb}{0.784, 0.003, 0.313}
\definecolor{pink}{cmyk}{0, 0.7808, 0.4429, 0.1412}
\definecolor{amethyst}{rgb}{0.6, 0.4, 0.8}
\definecolor{black}{rgb}{0.0, 0.0, 0.0}
\definecolor{tb3_yellow}{rgb}{0.996, 1.0, 0.6}
\definecolor{tb3_orange}{rgb}{0.980, 0.8, 0.604}
\definecolor{tb3_red}{rgb}{0.972, 0.6, 0.6}
\definecolor{dimgray}{rgb}{0.41, 0.41, 0.41}
\definecolor{brickred}{rgb}{0.8, 0.25, 0.33}
\definecolor{bleudefrance}{rgb}{0.19, 0.55, 0.91}
\definecolor{blue(ncs)}{rgb}{0.265, 0.445, 0.765}
\definecolor{blue(ryb)}{rgb}{0.01, 0.28, 1.0}
\definecolor{orange}{rgb}{1.0, 0.49, 0.0}
\definecolor{Gray}{gray}{0.88}
\definecolor{green(ncs)}{rgb}{0.0, 0.62, 0.42}
\definecolor{brightpink}{rgb}{1.0, 0.0, 0.5}

\definecolor{kellygreen}{rgb}{0.3, 0.73, 0.09}
\newcommand{\colorref}[1]{{\color{kellygreen}{#1}}}

\newcolumntype{g}{>{\columncolor{CuGray}}c}
\newcolumntype{z}{>{\columncolor{CuGray}}l}

\renewcommand{\paragraph}[1]{\vspace{1mm}\noindent\textbf{#1.}\,\,}
\newcommand{\oh}[1]{\textcolor{azure(colorwheel)}{#1}}

\newcommand{\yw}[1]{\textcolor{yw}{#1}}
\newcommand{\orange}[1]{\textcolor{orange}{#1}}

\newcommand{\blue}[1]{\textcolor{blue(ryb)}{#1}}
\newcommand{\blueyw}[1]{\textcolor{blue(ncs)}{#1}}

\newcommand{\greencap}[1]{\textcolor{kellygreen}{#1}}

\usepackage{xspace}

\newcommand*\samethanks[1][\value{footnote}]{\footnotemark[#1]}
\makeatletter
\def\@fnsymbol#1{\ensuremath{\ifcase#1\or *\or \dagger\or \ddagger\or
   \mathsection\or \mathparagraph\or \|\or **\or \dagger\dagger
   \or \ddagger\ddagger \else\@ctrerr\fi}}
\makeatother

\def\onedot{.\@\xspace}
\def\eg{\emph{e.g}\onedot} 
\def\ie{\emph{i.e}\onedot}

\def\wrt{\emph{w.r.t}\onedot} 
\def\etal{\emph{et al}\onedot}

\newcommand{\Sref}[1]{Sec.~\ref{#1}}
\newcommand{\Eref}[1]{Eq.~(\ref{#1})}
\newcommand{\Fref}[1]{Fig.~\ref{#1}}
\newcommand{\Tref}[1]{Table~\ref{#1}}

\newcommand{\bb}{{\mathbf{b}}}

\newcommand{\bm}{{\mathbf{m}}}

\newcommand{\bp}{{\mathbf{p}}}

\newcommand{\br}{{\mathbf{r}}}

\newcommand{\bw}{{\mathbf{w}}}
\newcommand{\bx}{{\mathbf{x}}}
\newcommand{\by}{{\mathbf{y}}}

\newcommand{\bW}{\mathbf{W}}

\newcommand{\calL}{{\mathcal{L}}}

\newcommand{\bbeta}{\mbox{\boldmath $\beta$}}

\newcommand{\btheta}{\mbox{\boldmath $\theta$}}

\newcommand{\bpsi}{\mbox{\boldmath $\psi$}}

\newcommand{\Real}{\mathbb R}

\newcommand{\be}{\begin{eqnarray}}
\newcommand{\ee}{\end{eqnarray}}
\newcommand{\bee}{\begin{eqnarray*}}
\newcommand{\eee}{\end{eqnarray*}}

\newcommand{\matrixb}{\left[ \begin{array}}
\newcommand{\matrixe}{\end{array} \right]}

\usepackage[pagebackref,breaklinks,colorlinks=True,citecolor=ashblue,linkcolor=cornellred,bookmarks=false]{hyperref}
\usepackage{url}
\usepackage{graphicx}
\usepackage{amsmath}
\usepackage{amssymb}
\usepackage{booktabs}
\usepackage{amsthm}

\newtheorem{proposition}{Proposition}

\title{A Large-Scale 3D Face Mesh Video Dataset \\
via Neural Re-parameterized Optimization}

\author{Kim Youwang${}^{1}$, Lee Hyun${}^{1}$\thanks{Equally contributed 2\emph{nd} authors.}, Kim Sung-Bin${}^{1}$\samethanks, Suekyeong Nam${}^{3}$, Janghoon Ju${}^{3}$, Tae-Hyun Oh${}^{1,2,4}$ \\
${}^{1}$Dept.~of Electrical Engineering, POSTECH $\quad{}^{2}$Grad.~School of AI, POSTECH, $\quad{}^{3}$Krafton Inc. \\ 
${}^{4}$Institute for Convergence Research and Education in Advanced Technology, Yonsei University 
}

\iclrfinalcopy 
\begin{document}

\maketitle

\vspace{-3mm}
\begin{abstract}
We propose NeuFace, a 3D face mesh pseudo
annotation method on videos via neural re-parameterized
optimization.
Despite the huge progress in 3D face reconstruction methods, 
generating reliable 3D face labels for in-the-wild dynamic videos 
remains challenging.
Using NeuFace optimization, we annotate the per-view/-frame accurate and consistent face meshes on large-scale face videos, called the NeuFace-dataset.
We investigate how neural re-parameterization
helps to reconstruct image-aligned 
facial details on 3D meshes
via gradient analysis.
By exploiting the naturalness and diversity of 3D faces in our dataset,
we demonstrate the usefulness
of our dataset for
3D face-related tasks: 
improving the reconstruction accuracy 
of 
an
existing 
3D face reconstruction model and learning 3D facial motion prior.
%
Code and datasets will be available at \url{https://neuface-dataset.github.io}.
\end{abstract}\vspace{-1.5mm}

\section{Introduction}
\label{sec:introrevised}

A comprehensive understanding of \emph{dynamic} 3D human 
faces has been a long-standing problem in
computer vision and graphics. 
%
Reconstructing and generating
dynamic 3D human faces
are key components for diverse tasks such as 
face recognition~\citep{weyrauch2004component, volker2003facerec}, 
face forgery detection~\citep{cozzolino2020idreveal, roessler2018faceforensics,roessler2019faceforensicspp}, 
video face editing~\citep{mbr_frf,kim2018deep,tewari2020pie}, 
facial motion or expression transfer~\citep{thies2015realtime,thies2016face,thies2018headon},
XR applications~\citep{EgoChat20,wang2021facevid2vid,richard2021meshtalk}, and human avatar generation~\citep{raj2020pva, Ma_2021_CVPR,youwang2022clipactor}.

Recent studies~\citep{wood2021fake,wood2022dense,bae2023digiface1m,yeh2022learning} have shown that reliable datasets of facial geometry, 
even synthetic or pseudo ones, 
can help achieve
a comprehensive understanding of 
\emph{``static''}
3D faces.
However, there is currently a lack of 
reliable and large-scale datasets 
containing
\emph{``dynamic''} and \emph{``natural''} 3D facial motion annotations.
The
lack of such
datasets
%
becomes a
bottleneck
for studying
inherent facial motion dynamics
or
3D face reconstruction tasks 
by restricting them to rely on weak supervision, \eg, 2D landmarks or segmentation maps.
%
%
%
%
Accurately acquired 3D face video data may mitigate such issues but
typically 
requires intensive and time-consuming efforts with 
carefully calibrated multi-view cameras and 
controlled 
lighting conditions~\citep{humbi2021tpami,joo_iccv_2015,totalcap_joo_2018,VOCA2019,COMA:ECCV18}.
Few seminal works~\citep{fanelli20103, COMA:ECCV18, VOCA2019, MICA:ECCV2022} 
take such effort to build 3D face video datasets.
Despite significant efforts, the existing datasets obtained from 
such restricted settings are limited in scale, scenarios, diversity of actor identity and expression,
and naturalness of facial motion (see \Tref{tab:data_stat}).

In contrast to 3D, 
there are 
an incomparably large amount of 2D face video datasets available online~\citep{kaisiyuan2020mead, Nagrani17,voxceleb2, zhu2022celebvhq, vggface, vggface2, stylegan, wang2021facevid2vid, Wang_2019_ICCV, liu2015faceattributes},
which are captured in diverse 
in-the-wild
environments but without 3D annotations.
%
%
%
%
As successfully demonstrated in some 3D tasks~\citep{fang2021mirrored,bouazizi2021learning,huang2022rich,mueller2021on,hassan2019prox,bayer2016on,ng2022learning2listen} as well as other analysis tasks~\citep{miech19howto100m,nagrani2022learning,lee2021acav100m},
leveraging off-the-shelf reconstruction models is a common practice to obtain pseudo ground-truth of such in-the-wild 
videos that were already captured.
They showed that high-quality and large-scale pseudo ground-truth is sufficient to achieve the state-of-the-art 
at the time of 
their works.
%
Similarly, a na\"ive approach is to 
construct a large-scale 3D face video dataset by curating existing 
2D video datasets
and obtain 3D face annotations with off-the-shelf face reconstruction models~\citep{DECA:Siggraph2021,EMOCA:CVPR:2022}.
%
However, existing
3D face reconstruction models
have limitations for 
reconstructing 
temporally smooth or multi-view consistent 3D face meshes from videos.
%
This is because 
state-of-the-art
face reconstruction models are
typically trained on single-view static images only with 2D supervision;
thus fail to extrapolate to 
faces 
having rare poses
and 
yield jittered motion due to the per-frame independent inference.
%
%

To address these difficulties, we propose \textbf{NeuFace optimization}, 
%
which reconstructs accurate and spatio-temporally consistent parametric 3D face meshes on videos.
By re-parameterizing 3D face meshes with
neural network parameters, 
NeuFace 
infuses spatio-temporal cues
of dynamic face videos on 3D face reconstruction.
NeuFace optimizes
spatio-temporal consistency
losses and 
the 2D landmark loss to acquire reliable face mesh pseudo-labels for videos. 

\definecolor{Gray}{gray}{0.88}
\begin{wraptable}{R}{0.55\textwidth} 
    \centering
    \vspace{-4mm}
    \resizebox{\linewidth}{!}{
    \begin{tabular}{l c@{\,\,\,}c@{\,\,\,\,}c@{\,\,\,\,}c@{\,\,}c}
    \toprule
    Dataset& No. seq. [K]&No. id&Dur. [hrs]&Env.\\
        
    \cmidrule{1-5}
    \multicolumn{4}{l}{\textbf{Existing 3D face video datasets}}\\
    BIWI 3D & 1.1 & 14 & 1.4 & Lab. \\
    COMA & 0.15 & 12 & 0.1 & Lab. \\
    VOCASET & 0.5 & 12 & 0.5 & Lab.\\
    \cmidrule{1-5}
    \rowcolor{Gray}
\textbf{NeuFace-dataset (ours)} & \textbf{1,245} & \textbf{21,048} & \textbf{2,090} & \textbf{Wild + Lab.}\\
\quad $\vdash$ NeuFace$_{\text{MEAD}}$ & 210 & 48 & 25 & Lab.\\
\quad $\vdash$ NeuFace$_{\text{VoxCeleb2}}$ & 1,000 & 6,000 & 2,000 & Wild\\
\quad $\vdash$ NeuFace$_{\text{CelebV-HQ}}$ & 35 & 15,000 & 65 & Wild\\
\bottomrule
\end{tabular}}
\caption{
    \textbf{NeuFace-dataset} provides reliable 3D face mesh annotations for MEAD, VoxCeleb2 and CelebV-HQ videos, 
    which is significantly richer
    than
    the existing 
    datasets in terms of the scale, diversity and naturalness. 
    $\emph{Abbr.}$ \{seq.: sequences, id.: identities, Dur.: duration, Env.: environment\}
    }
   \label{tab:data_stat}
\end{wraptable}

Using this method, we create the
\textbf{NeuFace-dataset}, the first large-scale, accurate and spatio-temporally consistent 3D face meshes for videos.
%
Our dataset 
contains 3D face mesh pseudo-labels for 
large-scale, multi-view or in-the-wild 2D face videos, 
MEAD~\citep{kaisiyuan2020mead}, 
VoxCeleb2~\citep{voxceleb2}, 
and CelebV-HQ~\citep{zhu2022celebvhq}, 
achieving about 1,000 times
larger number of sequences than 
existing facial
motion capture datasets
(see \Tref{tab:data_stat}). 
%
Our dataset 
inherits the benefits of
the rich visual attributes in
large-scale face videos, \eg, 
various races, appearances, backgrounds, natural facial motions, and expressions. 
We assess the fidelity
of our dataset
by investigating the 
cross-view vertex distance and the 3D motion stability index. 
%
We demonstrate that our dataset contains more 
spatio-temporally consistent and accurate 3D 
meshes than the 
competing datasets built with strong baseline methods.
To demonstrate the potential of our dataset, we present two applications: (1) improving the accuracy of a face reconstruction model and (2) learning a generative 3D facial motion prior.
%
These applications highlight that NeuFace-dataset
can be further used in 
diverse 
applications
demanding high-quality 
and large-scale 3D face meshes.
We summarize our main contributions as follows:
\begin{itemize}
    \item 
    \textbf{NeuFace}, an optimization method for reconstructing accurate and spatio-temporally consistent 3D face meshes on videos via neural re-parameterization.
    \item \textbf{NeuFace-dataset}, the first large-scale 
    3D face mesh pseudo-labels
    constructed by 
    curating
    existing large-scale 2D face video datasets with our method.
    \item 
    Demonstrating the benefits of NeuFace-dataset: (1) improve the accuracy of 
    off-the-shelf 
    face mesh regressors,
    (2) learn 3D facial motion prior for long-term face motion generation.
\end{itemize}

\section{Related work}

\paragraph{3D face datasets}
To achieve a comprehensive understanding of dynamic 3D faces, 
large-scale in-the-wild 3D face video datasets are essential.
%
%
There exist large-scale
2D face datasets that provide expressive
face images or videos~\citep{kaisiyuan2020mead, Nagrani17,voxceleb2, zhu2022celebvhq, vggface, vggface2, stylegan, liu2015faceattributes}
with diverse attributes covering a wide variety of appearances, races, environments, scenarios, and emotions.
However, most 2D face datasets do not have corresponding 3D annotations, 
due to the difficulty of 3D face 
acquisition, 
especially for in-the-wild environments.
%
Although some recent datasets~\citep{humbi2021tpami,COMA:ECCV18,VOCA2019,MICA:ECCV2022,wood2021fake} 
provide 3D face annotations with paired images or videos,%
\footnote{MICA released the medium-scale 3D annotated face datasets,
but only a single identity 
parameter
per video is provided, not the facial
poses or expression parameters, \ie, static 3D faces.}
they are acquired in the restricted and carefully controlled 
indoor capturing environment, \eg, laboratory,
yielding small scale, unnatural facial expressions
and a limited variety of facial identities or features.
Achieving in-the-wild naturalness and acquiring true 3D labels would be mutually exclusive in the real-world.
Due to the challenge of constructing a real-world 3D face dataset, FaceSynthetics~\citep{wood2021fake} synthesizes
large-scale synthetic face images and 
annotations derived from synthetic 3D faces, but limited in 
that they only publish
images
and 2D annotations without 3D annotations, which restrict 3D face video applications.%

%
%

%


In this work, we present the \textbf{NeuFace-dataset}, the first large-scale 
3D face mesh pseudo-labels
paired with the existing in-the-wild 2D face video datasets, 
resolving the lack of the 3D face video datasets. 


\paragraph{3D face reconstruction}
To obtain
reliable face meshes
for 
large-scale face videos, we need
accurate 3D face reconstruction methods for videos.
Reconstructing accurate 3D faces 
from limited visual cues, 
\eg, a monocular image, 
is an ill-posed problem.
Model-based approaches have been the mainstream to mitigate the ill-posedness
and 
have advanced 
with the 3D Morphable Models (3DMMs)~\citep{volker1999mmface, bfm09, FLAME:SiggraphAsia2017}
and 3DMM-based reconstruction methods~\citep{zollhoefer2018facestar,3dmm2020survey,DECA:Siggraph2021,EMOCA:CVPR:2022,MICA:ECCV2022}.
%
%

3D face reconstruction methods can be 
categorized into 
learning-based and optimization-based approaches. 
%
The learning-based approaches, \eg, \citep{DECA:Siggraph2021, EMOCA:CVPR:2022, MICA:ECCV2022, RingNet:CVPR:2019, tran2016regressing},
use
neural networks trained on 
large-scale face image datasets
to regress the 3DMM parameters from a single image. 
%
The optimization-based approaches~\citep{volker2003facerec, patrik2015fitting,Chen2013AccurateAR, wood2022dense, thies2015realtime, Gecer_2019_CVPR} 
optimize the 2D landmark or photometric losses with 
extra
regularization terms directly over the 3DMM parameters.
%
%
Given a 
specific
image, these methods overfit to 2D landmarks observations,
thus 
showing better 2D landmark fit than the learning-based methods.
These approaches are suitable for our purpose in that we need accurate reconstruction that best fits each video.
%
%
However, 
the regularization terms are typically hand-designed with prior assumptions that disregard the input image.
%
These regularization terms often introduce mean shape biases~\citep{DECA:Siggraph2021,SMPL-X:2019,bogo2016smplify,joo2020eft}, 
due to their independence to input data, which we call
the \emph{data-independent prior}.
Also, balancing the losses and regularization is inherently cumbersome and may introduce initialization sensitivity and local minima issues~\citep{joo2020eft,SMPL-X:2019,bogo2016smplify,choutas2020expose}.

Instead of hand-designed regularization terms, we induce 
such effects by optimizing re-parameterized
3DMM parameters with a 
3DMM regression 
neural network, called \textbf{NeuFace} optimization.
Such
network parameters are trained from large-scale 
real face images, which implicitly embed strong prior from the trained data.
Thereby, we can leverage the favorable properties of the neural re-parameterization: 1) an input \emph{data-dependent} initialization and prior in 3DMM parameter optimization, 2) less bias toward a mean shape, and 3) stable optimization robust to local minima by over-parameterized model~\citep{cooper2021global,du2019gradient,neyshabur2018towards,allenzhu2019convergence,du2019grad_iclr}.
A similar re-parameterization was proposed in \citep{joo2020eft}, but they focus on human body in a single image input with fixed 2D landmark supervision.
We extend it to dynamic faces in the multi-view and video settings by sharing the neural parameters across views and frames, and 
devise an 
alternating optimization to self-supervise spatio-temporal consistency.
%

%
%
%
%

%

\section{NeuFace: 
a 3D face mesh optimization for videos
via neural re-parameterization}
%

In this section, we 
introduce the
neural re-parameterization of 3DMM (\Sref{sec:reparam}) 
and NeuFace, 
an optimization to obtain
accurate and spatio-temporally consistent face meshes
from face videos
(\Sref{sec:NeuFace}). 
We 
discuss 
the benefit of
neural re-parameterization
(\Sref{sec:why_reparam}), and 
show the possibility of our system as a reliable face mesh annotator (\Sref{sec:voca_gt}).

%
\subsection{Neural re-parameterization of 3D face meshes}
\label{sec:reparam}
We use FLAME~\citep{FLAME:SiggraphAsia2017}, a renowned 3DMM, as a 3D face representation.
3D face mesh vertices $\mathbf{M}$ and 
facial landmarks $\mathbf{J}$ 
for $F$
frame videos 
can be acquired with the differentiable skinning:
$\mathbf{M}, \mathbf{J}{=}\texttt{FLAME}(\boldsymbol{\mathbf{r}, \boldsymbol{\theta}, \boldsymbol{\beta}, \boldsymbol{\psi}})$,
where 
$\mathbf{r}$, $\boldsymbol{\theta}$,
$\boldsymbol{\beta}$ and $\boldsymbol{\psi}$
denote the head orientation,
face poses, face shape and expression coefficients, respectively.
%
For simplicity, FLAME parameters $\boldsymbol{\Theta}$
can be represented as, $\boldsymbol{\Theta} = [\mathbf{r}, \boldsymbol{\theta}, \boldsymbol{\beta}, \boldsymbol{\psi}]$.
We further re-parameterize
the FLAME parameters 
$\boldsymbol{\Theta}$ and weak perspective camera parameters $\mathbf{p}\in\mathbb{R}^{F\times 3}$ for video frames $\{\mathbf{I}_{f}\}^{F}_{f=1}$, into a 
neural network, $\Phi$, with parameters $\textbf{w}$,
%
\ie, $[\boldsymbol{\Theta}, \mathbf{p}] = \Phi_{\mathbf{w}}(\{\mathbf{I}_{f}\}^{F}_{f=1})$. 
%
We use the pre-trained DECA~\citep{DECA:Siggraph2021} or EMOCA~\citep{EMOCA:CVPR:2022} encoder for $\Phi_{\mathbf{w}}$. 
%

\begin{figure*}[t]
  \centering
\includegraphics[width=\linewidth]{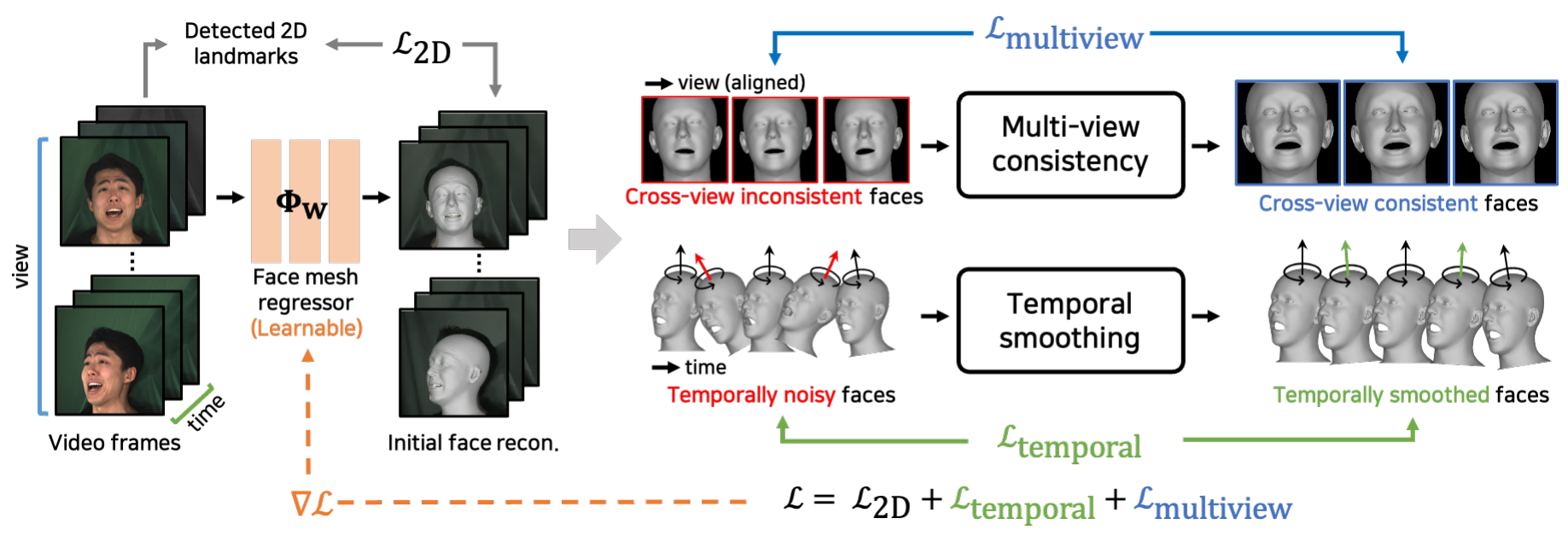}\vspace{-1.5mm}
   \caption{\textbf{NeuFace optimization.}
   Given 2D face
   videos, NeuFace optimizes spatio-temporally consistent 3D face meshes. 
   NeuFace updates the neural network parameters that
   re-parameterize the 3D face meshes with 
   2D landmark loss and spatio-temporal consistency losses.
   %
   %
   %
   %
   %
   }
   \label{fig:system}
\end{figure*}

\subsection{NeuFace optimization}
\label{sec:NeuFace}
Given the $N_{F}$ frames and $N_{V}$ views of a
face 
video 
$\{\mathbf{I}_{f,v}\}^{N_{F},N_{V}}_{f=1,v=1}$,
NeuFace 
aims to find the optimal
neural 
network
parameter $\mathbf{w}^{*}$ that
re-parameterizes accurate, multi-view and temporally consistent face meshes (see \Fref{fig:system}).
%
%
The optimization objective is defined as:
\begin{equation}
    \label{eq:obj}
    \mathbf{w}^{*} = \argmin_{\mathbf{w}} \,\, \mathcal{L}_{\textrm{2D}} + \lambda_{\textrm{temp}}\mathcal{L}_{\textrm{temporal}} + \lambda_{\textrm{view}}\mathcal{L}_{\textrm{multiview}},
\end{equation}
where 
$\{\lambda_{*}\}$ denotes the weights for each loss term.
%
%
%
Complex temporal and multi-view dependencies among 
variables in the losses would make direct optimization difficult~\citep{afonso2010augmented, salzmann2013continuous, zhang1993mean}. 
We ease the optimization of \Eref{eq:obj} by introducing latent target variables for self-supervision
in 
an Expectation-Maximization (EM) 
style optimization. 

\paragraph{2D landmark loss}
%
For 
each iteration $t$, we compute 
$\mathcal{L}_{\textrm{2D}}$ as a unary term, following the conventional
2D facial landmark re-projection loss~\citep{DECA:Siggraph2021,EMOCA:CVPR:2022} for the 
landmarks in all different 
frames and views: 
\begin{equation}
\label{eq:l_2d}
    \mathcal{L}_{\textrm{2D}} = {\frac{1}{N_{F}N_{V}}}\textstyle{\sum\nolimits_{f=1,v=1}^{N_{F},N_{V}}}
    {\lVert \pi(\mathbf{J}^{t}_{f,v}(\bw),\mathbf{p}^{t}_{f,v})- \mathbf{{j}}_{f,v} \rVert}_{1},
\end{equation}
where
$\pi(\cdot,\cdot)$ denotes the weak perspective projection, and 
$\mathbf{J}(\bw)$ 
is
the 
3D landmark from $\Phi_{\mathbf{w}}(\cdot)$.
\Eref{eq:l_2d} computes 
the 
pixel distance between 
the pre-detected 2D facial landmarks 
$\mathbf{{j}}$
and the regressed and projected 3D facial landmarks $\pi(\mathbf{J}(\bw), \mathbf{p})$.
%
$\mathbf{j}$ 
stays the same for
the whole optimization.
%

\paragraph{Temporal consistency loss}
Our temporal consistency loss 
reduces 
facial motion jitter caused by per-frame independent mesh 
regression on videos. 
Instead of 
a complicated Markov chain style loss, for each iteration $t$,
we first estimate
latent target meshes 
that represent temporally smooth heads in Expectation step (E-step).
Then, we simply maximize the likelihood of 
regressed meshes 
to its corresponding latent target in Maximization step (M-step).
In E-step, we feed $\{\mathbf{I}_{f,v}\}^{N_{F},N_{V}}_{f=1,v=1}$ into the network $\Phi_{\mathbf{w}^{t}}$ and obtain FLAME and camera parameters, $[\boldsymbol{\Theta}^{t}, \mathbf{p}^{t}]$.
For multiple
frames in 
view $v$, we extract the head orientations $\mathbf{r}^{t}_{:,v}$,
from 
$\boldsymbol{\Theta}^{t}$ and convert it to the unit quaternion $\mathbf{q}^{t}_{:,v}$.
To generate the latent target, \ie, temporally smooth
head orientations $\mathbf{\hat{q}}^{t}_{:, v}$,
we take the 
temporal moving average 
over
$\mathbf{q}^{t}_{:,v}$.
%
In M-step, we compute the temporal consistency loss
as:
\begin{equation}
\label{eq:l_temp}
    \mathcal{L}_{\textrm{temporal}} = \frac{1}{N_{F}N_{V}} \textstyle{\sum\nolimits_{f=1,v=1}^{N_{F},N_{V}}}{\lVert \mathbf{q}^{t}_{f,v}-\mathbf{\hat{q}}^{t}_{f,v} \rVert}_{2},
\end{equation}
where $\mathbf{q}$ is
the unit-quaternion representation of 
$\mathbf{r}$.
%
We empirically found that such simple consistency loss is sufficient enough to obtain temporal smoothness while allowing more flexible expressions.

\begin{wrapfigure}{r}{0.5\linewidth}
    \centering
    \vspace{-3mm}
    \includegraphics[width=1\linewidth]{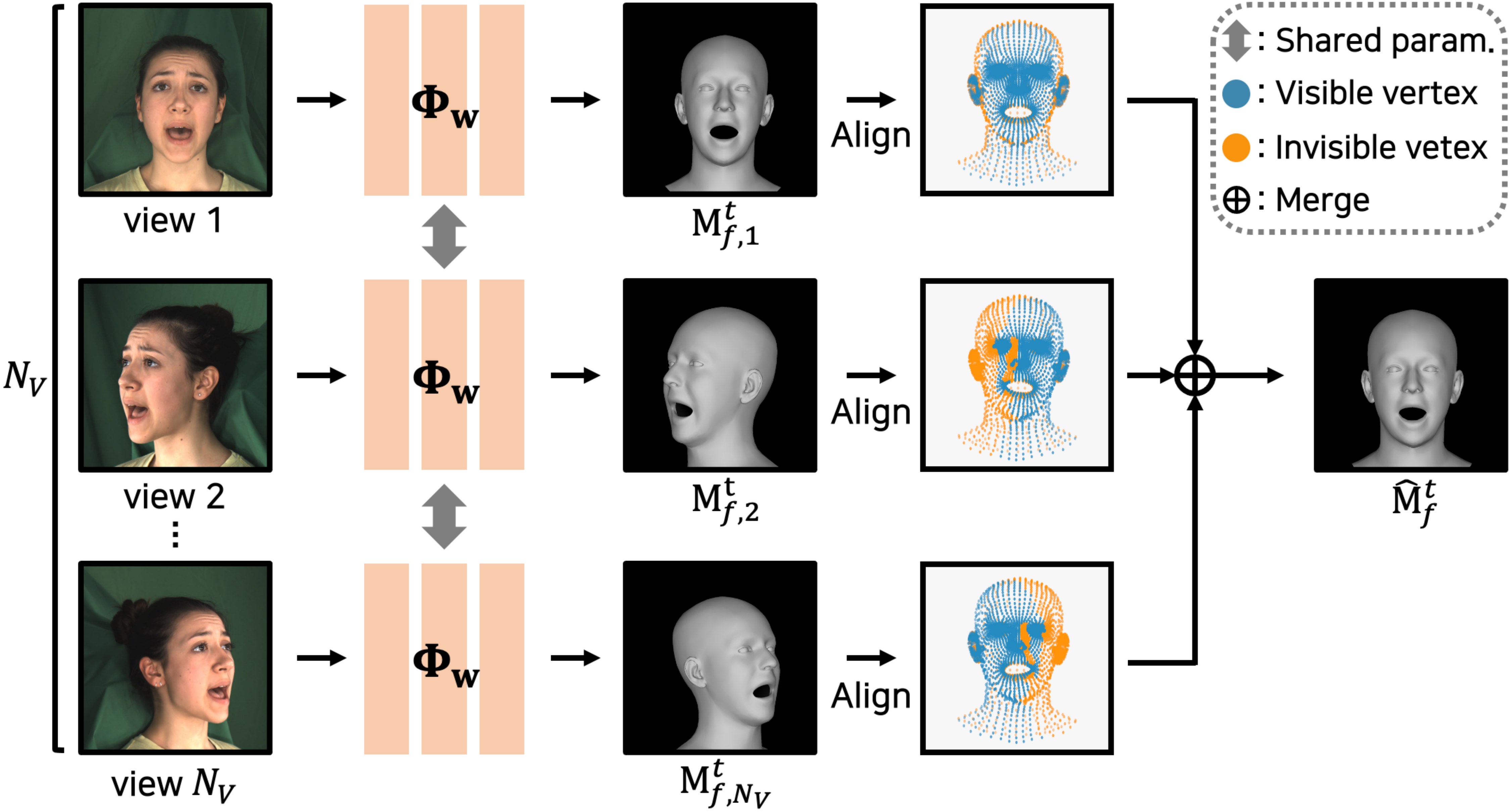}\vspace{-2mm}
    \caption{\textbf{Multi-view bootstrapping.}
    %
    Given initial mesh predictions for each view in frame $f$, 
    %
    we align 
    and merge the meshes depending on the confidence. The boostrapped mesh serves as a target for computing $\mathcal{L}_\text{multiview}$.
    }
    \vspace{-3mm}
    \label{fig:multiview}
\end{wrapfigure}

\paragraph{Multi-view consistency loss}
%
Although the aforementioned $\mathcal{L}_\text{2D}$
roughly guides the multi-view consistency of landmarks,
it cannot guarantee the consistency for off-landmark or invisible facial regions 
across views.
Therefore,
for multi-view captured face videos~\citep{kaisiyuan2020mead}, we leverage a simple principle to obtain 
consistent meshes over different views: face geometry should be consistent across views at the same time.
%
The goal 
is to 
bootstrap the per-view estimated noisy meshes 
by referencing the visible, or highly confident facial regions across different views.
Analogous to the temporal consistency loss, 
in M-step, we compute
the multi-view consistency loss as follows:
\begin{equation}
\label{eq:l_mv}
    \mathcal{L}_{\textrm{multiview}} = 
    \frac{1}{N_{F}N_{V}} 
    \textstyle{\sum\nolimits_{f=1, v=1}^{N_{F}, N_{V}}}
    {{\lVert \mathbf{M}^{t}_{f,v}-\mathbf{\hat{M}}^{t}_f \rVert_{1}}},
\end{equation}
where $\mathbf{\hat{M}}^{t}_f$ denotes the latent target mesh vertices estimated in E-step of each iteration.
In E-step,
given vertices $\mathbf{M}^{t}_{f,:}$ of multiple views in frame $f$,
%
we interpret the vertex visibility as the per-vertex confidence.
%
To obtain the latent target mesh $\mathbf{\hat{M}}^{t}_f$,
we 
align per-view estimated meshes
to the canonical view,
and bootstrap the
meshes by 
taking the weighted average of $\mathbf{M}^{t}_{f,:}$
depending on the confidence (see \Fref{fig:multiview}).
%
With this, \Eref{eq:l_mv}
constrains the vertices of each view to be
consistent with $\mathbf{\hat{M}}^{t}_f$.



\paragraph{Overall process}
We first estimate all the latent variables, $\mathbf{\hat{q}}$ and $\mathbf{\hat{M}}$ as E-step.
With the estimated latent variables as the self-supervision target, 
we optimize \Eref{eq:obj} over 
the \emph{network parameter} $\mathbf{w}$ as M-step.
This single alternating iteration
updates the optimization parameter $\mathbf{w}^t{\rightarrow}\mathbf{w}^{t+1}$ at iteration $t$.
We iterate alternating E-step and M-step until convergence.
After
convergence, we obtain the final solution
$[\boldsymbol{\Theta}^{*}, \mathbf{p}^{*}]$
by querying video frames to the optimized network, \ie, $[\boldsymbol{\Theta}^{*}, \mathbf{p}^{*}]{=}\Phi_{\mathbf{w}^{*}}(\{\mathbf{I}_{f,v}\}^{N_{F},N_{V}}_{f=1,v=1})$.

\subsection{Why is NeuFace optimization effective?}
\label{sec:why_reparam}
Note that one can simply update
FLAME parameters directly with the same loss in \Eref{eq:obj}.
%
Then, why do we 
need neural re-parameterization of
3D face meshes?
%
We claim such neural re-parameterization 
allows 
\emph{data-dependent mesh update}, 
which 
the FLAME fitting cannot achieve.
To support our claim, we 
analyze the benefit of our 
optimization by comparing it with
the solid baseline.

\paragraph{Baseline: FLAME fitting}
%
%
%
Given the same
video frames $\{\mathbf{I}_{f,v}\}^{N_{F},N_{V}}_{f=1,v=1}$
and the same initial FLAME and camera parameters $[\boldsymbol{\Theta}_{\bb}, \mathbf{p}_{\bb}]$ as NeuFace\footnote{To conduct a fair comparison with a strong baseline, we initialize $[\boldsymbol{\Theta}_{\bb}, \mathbf{p}_{\bb}]$ as the prediction of the pre-trained DECA. This is identical to NeuFace optimization (\Eref{eq:obj}); only the optimization variable is different.},
%
we implement the baseline optimization as:
\begin{equation}
    \label{eq:baseline}
    [\boldsymbol{\Theta_{\bb}^{*}}, \mathbf{p}_{\bb}^{*}] = 
    \argmin_{\boldsymbol{\Theta_{\bb}, \mathbf{p}_{\bb}}} \,\,
    \mathcal{L}_{\textrm{2D}} + \lambda_{\textrm{temp}}\mathcal{L}_{\textrm{temporal}}
    + \lambda_{\textrm{view}}\mathcal{L}_{\textrm{multiview}}
    + \lambda_{\mathbf{r}}\mathcal{L}_{\mathbf{r}}
    + \lambda_{\boldsymbol{\theta}}\mathcal{L}_{\boldsymbol{\theta}}
    + \lambda_{\boldsymbol{\beta}}\mathcal{L}_{\boldsymbol{\beta}}
    + \lambda_{\boldsymbol{\psi}}\mathcal{L}_{\boldsymbol{\psi}},
\end{equation}
where the losses $\mathcal{L}_{\textrm{2D}}$, 
$\mathcal{L}_{\textrm{temporal}}$ and 
$\mathcal{L}_{\textrm{multiview}}$ 
are identical to the Eqs.~(\ref{eq:l_2d}), (\ref{eq:l_temp}), and (\ref{eq:l_mv}).
%
$\mathcal{L}_{\mathbf{r}}$,
$\mathcal{L}_{\boldsymbol{\theta}}$,
$\mathcal{L}_{\boldsymbol{\beta}}$ and
$\mathcal{L}_{\boldsymbol{\psi}}$,
are the common
regularization terms 
used in~\citep{FLAME:SiggraphAsia2017,wood2022dense}.

\paragraph{Data-dependent gradients for mesh update}
We analyze the data-dependency of the baseline and NeuFace optimization by investigating back-propagated gradients.
%
For the FLAME fitting (\Eref{eq:baseline}), the update rule for FLAME parameters $\boldsymbol{\Theta}_{\bb}$ at optimization step $t$ is as follows:
%
%
\begin{equation}
    \label{eq:flame_update}
    \boldsymbol{\Theta}_{\bb}^{t+1} = \boldsymbol{\Theta}_{\bb}^{t} -\alpha\frac{\partial\mathcal{L}}{\partial\boldsymbol{\Theta}_{\bb}^{t}},
\end{equation}
where $\mathcal{L}$ denotes the sum
of all the losses used in the optimization.
%
In contrast, given video frames $\{\mathbf{I}_{f,v}\}^{N_{F},N_{V}}_{f=1,v=1}$, or simply $\blue{\mathbf{I}}$, 
the update 
for our NeuFace optimization is as follows: 
\begin{equation}
    \label{eq:ours_update}
    \mathbf{w}^{t+1} = \mathbf{w}^{t} - \alpha\frac{\partial\mathcal{L}}{\partial\mathbf{w}^t} = \mathbf{w}^{t} - \alpha(
    \frac{\partial\mathcal{L}}{\partial\boldsymbol{\Theta}_{\bw}^t}\cdot\frac{\partial\boldsymbol{\Theta}_{\bw}^{t}}{\partial\mathbf{w}^t}
    ) = \mathbf{w}^{t} - \alpha(
    \frac{\partial\mathcal{L}}{\partial\boldsymbol{\Theta}_{\bw}^t}\cdot\frac{\partial}{\partial\mathbf{w}^t}\Phi_{\mathbf{w}^{t}}(\blue{\mathbf{I}})
    ),
\end{equation}
where $\boldsymbol{\Theta}_{\bw}^{t}$ is re-/over-parameterized 
by
the neural network $\Phi_{\mathbf{w}^{t}}$, 
%
\ie, $\boldsymbol{\Theta}_{\bw}^{t}=\Phi_{\mathbf{w}^{t}}(\blue{\mathbf{I}})$.
%

By comparing
the back-propagated gradient terms in Eqs.~(\ref{eq:flame_update}) and (\ref{eq:ours_update}),
%
%
we can intuitively notice that the update for NeuFace optimization (\Eref{eq:ours_update}) 
is conditioned 
by input $\blue{\mathbf{I}}$, yielding \emph{data-dependent} 
mesh update.
%
%
With
data-dependent gradient $\tfrac{\partial}{\partial\mathbf{w}^t}\Phi_{\mathbf{w}^{t}}(\blue{\mathbf{I}})$, 
NeuFace optimization may inherit the implicit prior
embedded in 
the pre-trained neural model, \eg, DECA~\citep{DECA:Siggraph2021}, learned from large-scale real face images.
This allows NeuFace optimization to obtain expressive and image-aligned facial details on meshes. 

It is also worthwhile to note that, thanks to over-parameterization of $\Phi_{\mathbf{w}}(\cdot)$ \wrt $\boldsymbol{\Theta}$, we benefit from the following favorable property. For simplicity, we consider a simple $l_2$-loss
and a 
fully connected ReLU network,\footnote{By sacrificing the complexity of proof, the same conclusion holds for  ResNet~\citep{allenzhu2019convergence}.} but it is sufficient to understand the 
mechanism
of NeuFace
optimization.

\begin{proposition}[Informal] \textbf{\emph{Global convergence.}}\label{thm:globaloptima}
For the input data $\mathbf{x}\in[0,1]^{n\times d_{in}}$, paired labels $\mathbf{y}^* \in \R^{n\times d_{out}}$, and an over-parameterized $L$-layer 
fully connected network $\Phi_{\mathbf{w}}(\cdot)$ with ReLU activation and uniform weight widths,
consider optimizing the non-convex problem: $\argmin_\mathbf{w}\mathcal{L}(\mathbf{w}) = \tfrac{1}{2}\|\Phi_{\mathbf{w}}(\mathbf{x}) - \mathbf{y}^*\|_2^2$.
Under some assumptions, gradient descent finds a global optimum in polynomial time with high probability.
\end{proposition}

\begin{wrapfigure}{r}{0.5\linewidth}
    \centering
    \vspace{-5mm}
    \includegraphics[width=\linewidth]{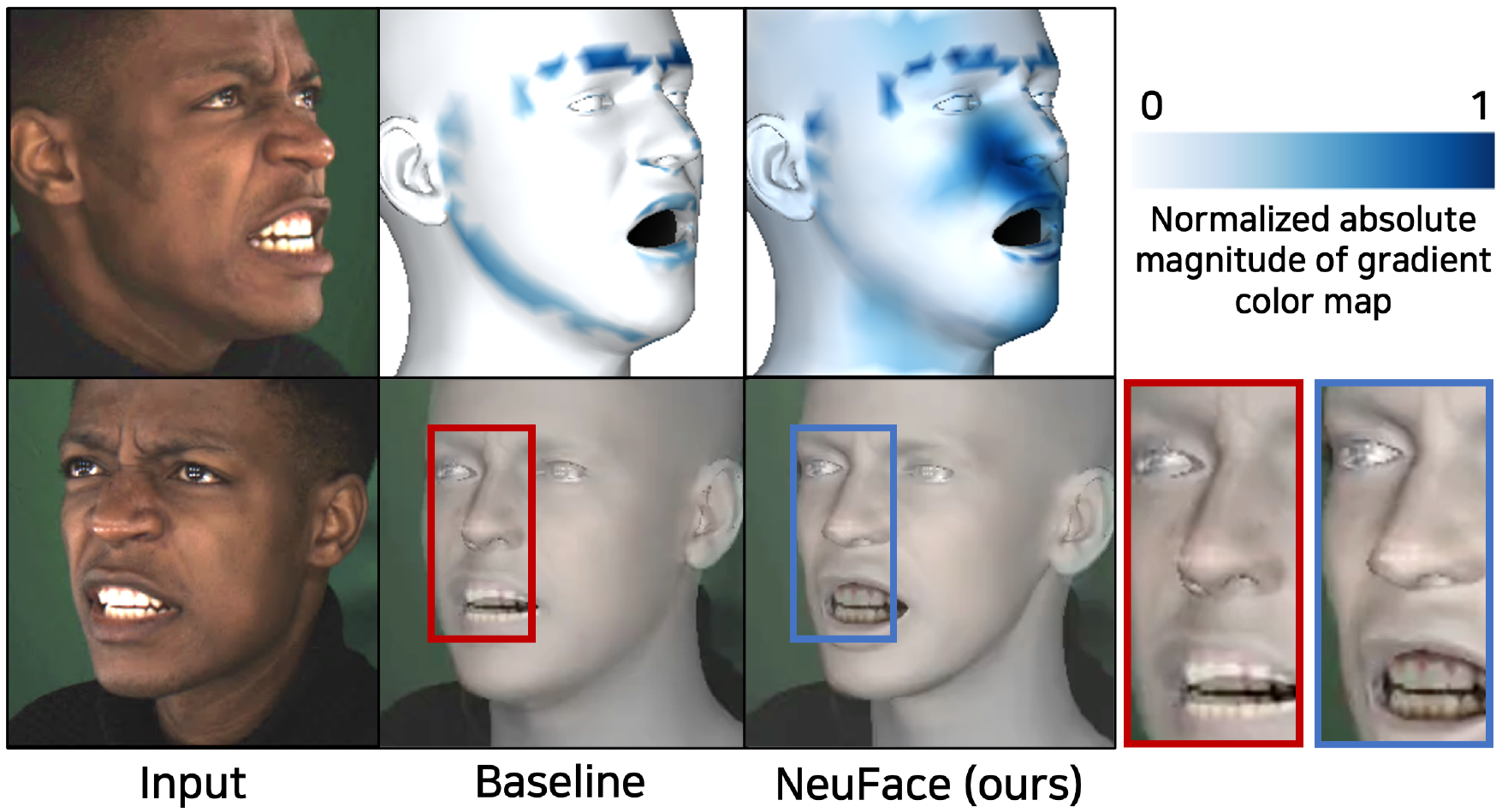}\vspace{-1.5mm}
    \caption{\textbf{Data-dependent gradient.}
    %
    NeuFace optimization obtains a richer gradient map regarding the pixel-level facial details
    (1\emph{st} row). 
    Thus,
    our method achieves more expressive and accurately image-aligned meshes than the baseline
    (2\emph{nd} row).}\vspace{-3mm}
    \label{fig:grad}
\end{wrapfigure}

\noindent Proposition~\ref{thm:globaloptima} can be derived by simply re-compositing the results by \citet{allenzhu2019convergence}.
Its proof sketch can be found in the supplementary material.
This hints that our over-parameterization helps NeuFace optimization achieve robustness to local minima
and avoid mean shape biases.


To see how data-dependent gradient of NeuFace affects the mesh optimization,
we visualize the absolute magnitude
of the back-propagated gradients of each method
in \Fref{fig:grad}.
The baseline optimization produces a sparse
gradient map 
along the face landmarks, which disregards 
the pixel-level facial details, \eg, wrinkles or facial boundaries.
%
In contrast, NeuFace additionally induces the dense gradients over face surfaces,
not just sparse landmarks,
%
which are helpful 
for representing 
image-aligned and detailed
facial expressions on meshes.
%
%
%
%
Thanks to the rich 
gradient map, 
our method yields more 
expressive and accurately image-aligned 
meshes than the baseline.
%




%
%


\subsection{How reliable is NeuFace optimization?}\label{sec:voca_gt}

\begin{wrapfigure}{r}{0.5\linewidth}
    \vspace{-5.5mm}
    \centering
    \includegraphics[width=\linewidth]{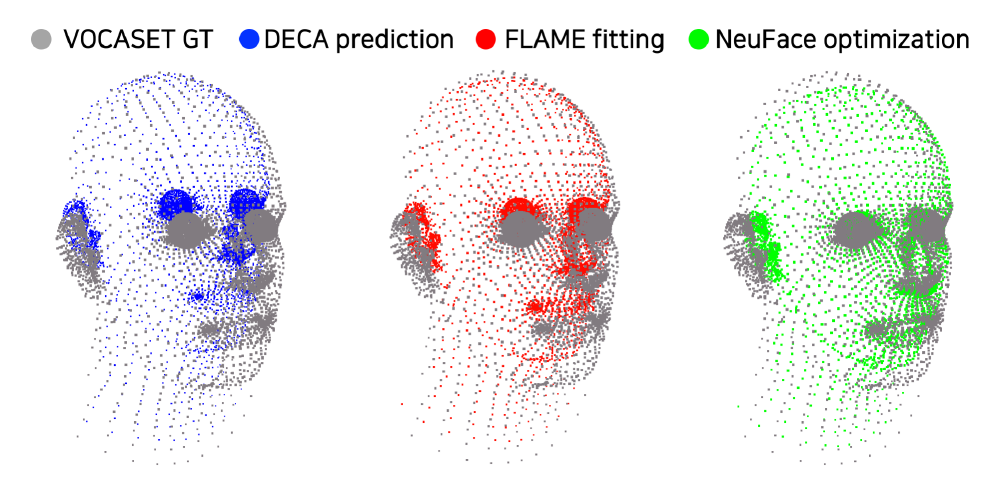}\vspace{-1mm}
  \resizebox{\linewidth}{!}{
  \begin{tabular}{l ccc}
  \toprule
  Method & DECA & FLAME fitting (\Eref{eq:baseline})& NeuFace (ours) \\
  \cmidrule(lr){1-4} 
  MPVE [\emph{mm}] $\downarrow$ & 38.4 & 31.5 & \textbf{30.6} \\
  \bottomrule
  \end{tabular}}\vspace{-2mm}
  \caption{\textbf{Accuracy 
  on MoCap dataset.}
  Given the ground-truth MoCap meshes,
  NeuFace optimization reconstructs more vertex-level accurate meshes than the competing methods.}
    \label{fig:vocaset}
\end{wrapfigure}

Many recent face-related applications~\citep{ng2022learning2listen,Khakhulin2022ROME,Feng:TRUST:ECCV2022} 
utilize a pre-trained, off-the-shelf 3D face reconstruction model or the FLAME fitting (\Eref{eq:baseline}) as a pseudo ground-truth annotator.
%
%
%
Compared to such conventional face mesh annotation methods, we discuss how reliable Neuface optimization is.
Specifically, we measure the vertex-level accuracy of the reconstructed face meshes by NeuFace optimization on the motion capture videos, VOCASET~\citep{VOCA2019}.

VOCASET is a small-scale facial motion capture dataset
that provides
registered ground-truth mesh sequences.
%
%
%
%
Given the ground-truth 
mesh sequences from the VOCASET,
we 
evaluate the 
Mean-Per-Vertex-Error (MPVE)~\citep{cho2022FastMETRO,lin2021endtoend,lin2021-mesh-graphormer} 
of face meshes obtained by 
pre-trained DECA, FLAME fitting and our method. 
%
%
%
%
In \Fref{fig:vocaset}, NeuFace optimization achieves more vertex-level accurate meshes than other methods, \ie,
lower MPVE.
Note that FLAME fitting still achieves competitive MPVE with ours, which shows that it is a valid, strong baseline.
%
%
Such favorable mesh accuracy of NeuFace optimization motivates us to leverage it 
%
%
as a reliable face mesh annotator for large-scale face videos, and build the NeuFace-dataset.
%



\begin{table*}[t!]
  \label{tab:quant}
  \centering
  \resizebox{\linewidth}{!}{
  \begin{tabular}{l cccc ccc ccc}
  \toprule
&\multicolumn{4}{c}{MEAD} & \multicolumn{3}{c}{VoxCeleb2} & \multicolumn{3}{c}{CelebV-HQ} \\
    \cmidrule(lr){2-5} \cmidrule(lr){6-8} \cmidrule(lr){9-11}
    \multicolumn{1}{c}{Dataset}  & \multicolumn{1}{c}{$\text{MSI}_\text{3D}^{\text{L}}\uparrow$} &\multicolumn{1}{c}{$\text{MSI}_\text{3D}^{\text{V}}\uparrow$} & \multicolumn{1}{c}{CVD $\downarrow$} & \multicolumn{1}{c}{NME $\downarrow$} & \multicolumn{1}{c}{$\text{MSI}_\text{3D}^{\text{L}}\uparrow$} &\multicolumn{1}{c}{$\text{MSI}_\text{3D}^{\text{V}}\uparrow$} & \multicolumn{1}{c}{NME $\downarrow$} & \multicolumn{1}{c}{$\text{MSI}_\text{3D}^{\text{L}}\uparrow$} &\multicolumn{1}{c}{$\text{MSI}_\text{3D}^{\text{V}}\uparrow$} & \multicolumn{1}{c}{NME $\downarrow$} \\
  \cmidrule{1-11}
  \multicolumn{1}{l}{Base-dataset (\Eref{eq:baseline})} & 0.034& 0.053 & 0.192 & 4.34 & 0.034 & 0.056 & 3.32 & 0.030 & 0.047 & 3.65\\
  \cmidrule{1-11}
\multicolumn{1}{l}{DECA-dataset}   & 0.011&0.016& 0.209 & 4.65 & 0.028 & 0.044 & 4.78 & 0.012 & 0.018 &5.34\\
\multicolumn{1}{l}{\textbf{NeuFace-D-dataset} (ours)}  & \textbf{0.206} & \textbf{0.305} & \textbf{0.103} & \textbf{2.58} & \textbf{0.095} & \textbf{0.137} & \textbf{2.19} & \textbf{0.054}&\textbf{0.074}&\textbf{2.55}\\
    \cmidrule{1-11}
    \multicolumn{1}{l}{EMOCA-dataset}   & 0.010 &0.016 & 0.199 & 5.42 & 0.003 & 0.004 & 4.77 & 0.005 & 0.007 &5.57\\
\multicolumn{1}{l}{\textbf{NeuFace-E-dataset} (ours)}  & \textbf{0.209} & \textbf{0.312} & \textbf{0.104} & \textbf{2.28} & \textbf{0.028} & \textbf{0.048} & \textbf{2.38} & \textbf{0.053}&\textbf{0.077}&\textbf{2.86}\\
    \bottomrule
  \end{tabular}}
  \caption{\textbf{Quantitative evaluation.}
  NeuFace-D/E-datasets (ours) 
  significantly outperform the other datasets in multi-view consistency (CVD), temporal consistency (MSI\textsubscript{3D}), and the 2D landmark accuracy (NME).
  $\emph{Abbr.}$ \{L: landmark, V: vertex.\}}\vspace{-1.5mm}
\label{evaluation_results}
\end{table*}

\begin{figure*}[t]
  \centering
  \includegraphics[width=0.8\linewidth]{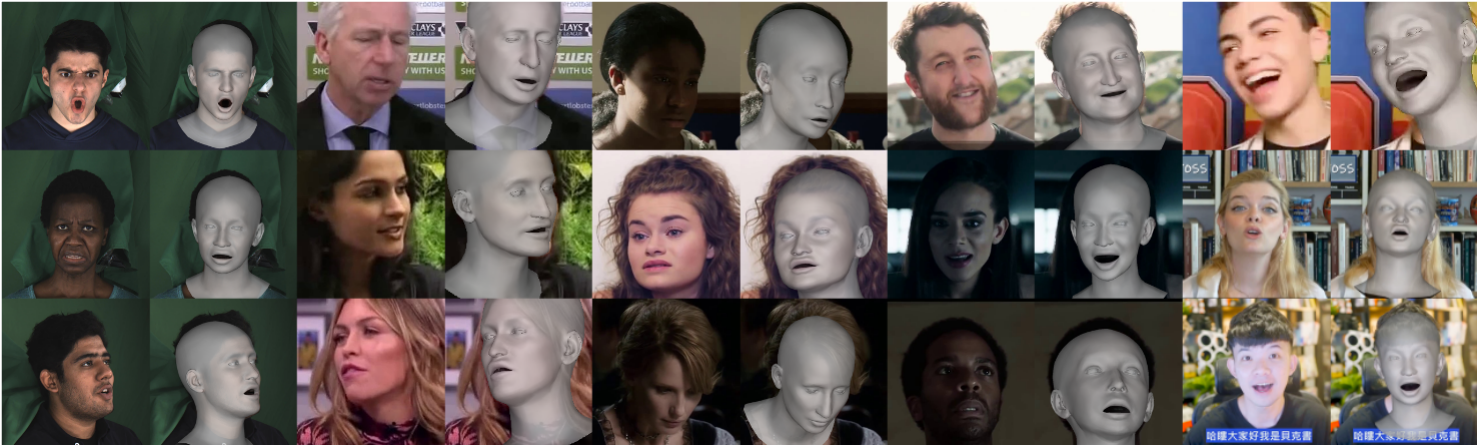}\vspace{-1.5mm}
   \caption{\textbf{NeuFace-dataset}
   %
   %
   contains accurate and spatio-temporally consistent 
   3D face mesh pseudo-labels
   for 
   large-scale video datasets.
   %
   %
   %
   Please refer to our \href{https://neuface-dataset.github.io}{project page}
   for more samples in \emph{video}.
   }
   \label{fig:teaser_data}
\end{figure*}

\section{The NeuFace-dataset}
%

\label{sec: dataset}
The NeuFace-dataset provides accurate and spatio-temporally consistent 
face meshes of
existing large-scale 2D face video datasets;
MEAD~\citep{kaisiyuan2020mead}, VoxCeleb2~\citep{voxceleb2}, and CelebV-HQ~\citep{zhu2022celebvhq} (see \Fref{fig:teaser_data}). 
Our datasets are denoted with 
NeuFace\textsubscript{$\{*\}$}
and summarized in \Tref{tab:data_stat}.
The NeuFace-dataset is, namely, the largest 3D face mesh pseudo-labeled dataset
in terms of the scale, naturalness, and diversity of facial attributes, emotions, and backgrounds. 
%
Please refer to the supplementary material for the dataset acquisition and filtering details.




We assess the fidelity of our dataset in terms of 
spatio-temporal consistency and landmark accuracy. 
%
We make competing datasets 
and compare the quality of the generated mesh annotations.
First, we compose the strong baseline, Base-dataset, by fitting FLAME with \Eref{eq:baseline}. 
%
We also utilize pre-trained DECA and EMOCA as mesh annotators and built DECA-dataset and EMOCA-dataset, respecitvely.
Finally, we build two versions of our dataset, \ie, NeuFace-D, and NeuFace-E, where each dataset is generated via \Eref{eq:obj} with DECA and EMOCA for the neural re-parameterization $\Phi_{\mathbf{w}}$, respectively.

\paragraph{Temporal consistency}
We extend the Motion Stability Index
(MSI)~\citep{ling2022stableface} to $\text{MSI}_\text{3D}$
and evaluate the temporal consistency of each dataset.
%
MSI\textsubscript{3D} computes a reciprocal of the motion 
acceleration variance of either 3D landmarks or 
vertices
and 
quantifies facial motion 
stability 
for a given $N_{F}$ frame video, $\{\mathbf{I}_{f}\}^{N_{F}}_{f=1}$,
as 
$\text{MSI}_\text{3D}(\{\mathbf{I}_{f}\}^{N_{F}}_{f=1}){=}\tfrac{1}{K}\sum_{i}\tfrac{1}{\sigma(\mathbf{a}^i)}$,
%
%
where $\mathbf{a}^i$ denotes the 3D motion acceleration of $i$-th 3D landmarks or vertices, 
$\sigma(\cdot)$ 
the temporal variance, 
and $K$ the number of 
landmarks or vertices.
%
If the mesh sequence has 
small temporal jittering, 
\ie, low motion variance,
it has a high $\text{MSI}_\text{3D}$ value. 
%
%
We 
compute $\text{MSI}_\text{3D}$ for landmarks and vertices, \ie, $\text{MSI}_\text{3D}^{\text{L}}$ and $\text{MSI}_\text{3D}^{\text{V}}$, respectively.
%
\Tref{evaluation_results} shows the 
$\text{MSI}_\text{3D}^{\text{L}}$ and $\text{MSI}_\text{3D}^{\text{V}}$
averaged over the validation sets. 
For the VoxCeleb2 and CelebV-HQ splits, the NeuFace-D/E-dataset outperform 
the 
other datasets
in 
both MSI\textsubscript{3D}s.
%
Remarkably, we have 
improvements on MSI\textsubscript{3D} 
more than 20 times in MEAD.
We postulate that the multi-view consistency loss 
also strengthens the temporal consistency for MEAD.
In other words, our 
losses would be mutually helpful when jointly optimized.
%
We discuss it through loss ablation studies in the supplementary material.


\begin{wrapfigure}{r}{0.5\linewidth}
    \vspace{-5mm}
    \centering
    \includegraphics[width=\linewidth]{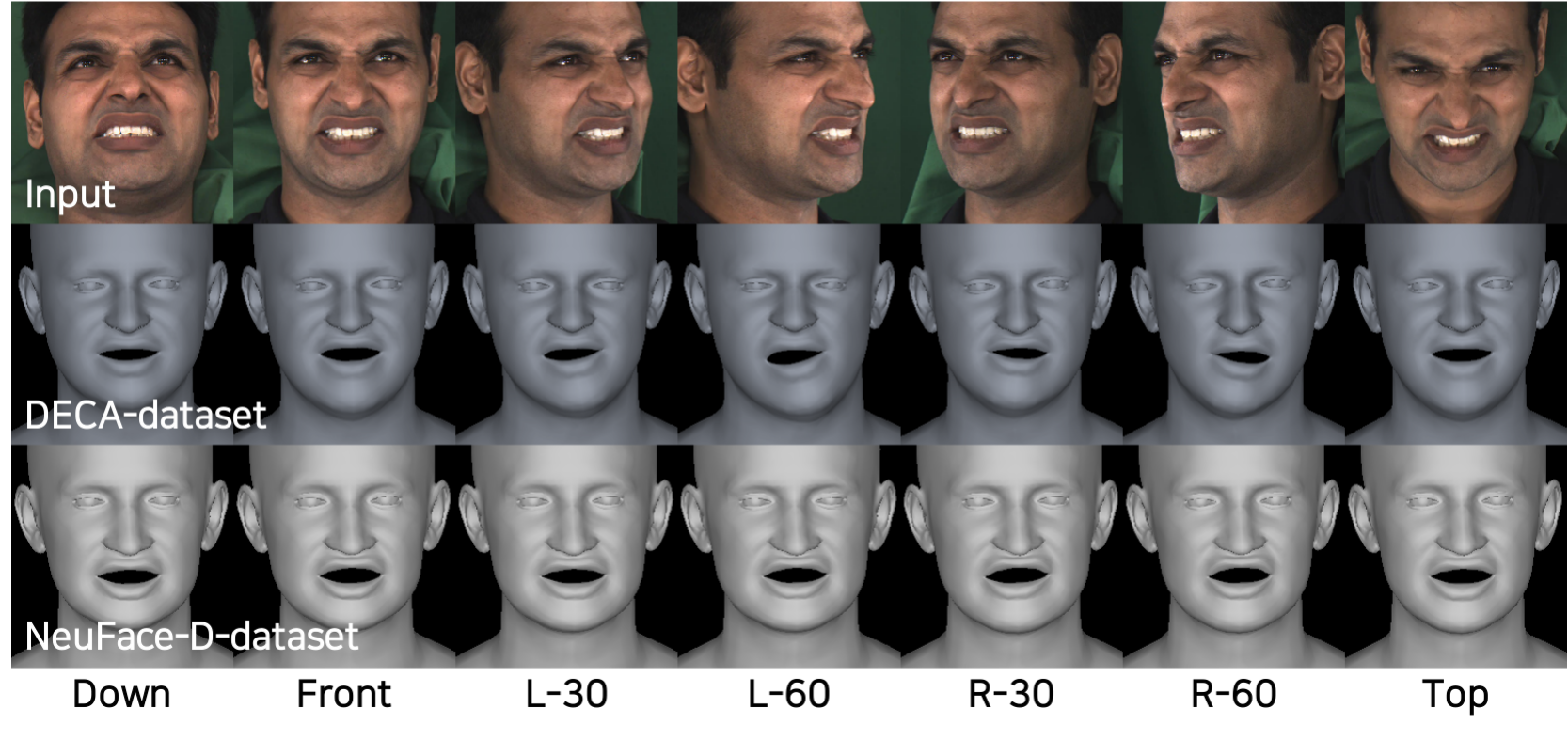}\vspace{-3mm}
    \caption{\textbf{Multi-view consistent face meshes.} 
    NeuFace-dataset contains multi-view consistent meshes compared to the DECA-dataset. L- and R- denote Left and Right, and 30 and 60 denote the camera view angles from the center.}\vspace{-3mm}
    \label{fig:spatial_qual}
\end{wrapfigure}


\paragraph{Multi-view consistency}
We visualize the predicted
meshes over different views
in 
\Fref{fig:spatial_qual}, where 
per-view independent estimations
are presented, not a single merged one.
We
verify that the NeuFace-D-dataset contains multi-view consistent meshes compared to the DECA-dataset, especially near the mouth region.
See supplementary material for the comparison of the EMOCA-dataset and NeuFace-E-dataset. 
%
As a quantitative
measure,
we 
compute the cross-view vertex distance (CVD), \ie, the vertex distance between two different views, $i$ and $j$, in the same frame $f$: ${\lVert \mathbf{M}_{f,i}{-}\mathbf{M}_{f,j} \rVert}_{1}$. 
We compare the averaged CVD of all views in \Tref{evaluation_results}.
CVD is only evaluated on the 
MEAD dataset, which is in a multi-camera setup.
While the DECA-/EMOCA-dataset results in high CVD, the NeuFace-dataset shows significantly lower CVD on overall views.

\paragraph{2D landmark accuracy}
A trivial solution to obtain
low CVD and high $\text{MSI}_{\text{3D}}$ is to 
regress the same mean face meshes 
across 
views and frames regardless of the input image.
To verify such occurrence, we 
measure the landmark accuracy of the
regressed 2D 
facial
landmarks 
using the 
normalized mean error (NME)~\citep{sagonas2016300}.
The NeuFace-D/E-dataset outperform the 
other datasets in NME, \ie, contain spatio-temporally consistent and accurately landmark-aligned meshes.

\section{Applications of the NeuFace-datasets}
In this section, we demonstrate the 
usefulness of the NeuFace-dataset.
%
%
%
We boost the
accuracy
of an off-the-shelf
face mesh regressor
by exploiting our dataset's
3D supervision
(\Sref{sec:retrain}).
Also, we 
%
learn generative facial motion prior 
from the large-scale, in-the-wild 3D faces in our dataset (\Sref{sec:humor}).

\subsection{Improving the 3D reconstruction accuracy}
%
\label{sec:retrain}
%
%
Due to the absence of large-scale 3D face video datasets,
%
existing face mesh regressor models utilize limited visual cues, such as 2D landmarks or segmentations.
Thus, we utilize the NeuFace-dataset to add direct 3D supervision to 
enhance the performance of such a model.

\paragraph{3D supervision with the NeuFace-dataset}
We implement the auxiliary 3D supervision as conventional 3D vertex and landmark losses~\citep{kolotouros2019spin,cho2022FastMETRO,lin2021endtoend,lin2021-mesh-graphormer}.
Given regressed and our annotated mesh vertices, ${\mathbf{M}}, \hat{\mathbf{M}}\in \mathbb{R}^{N_{M}\times3}$, 
and regressed and our annotated 3D landmarks, ${\mathbf{J}}, \hat{\mathbf{J}} \in\mathbb{R}^{N_{J}\times3}$,
the auxiliary 3D losses are
computed as:
$\mathcal{L}^{\text{M}}_\text{3D}{=}\tfrac{1}{N_{M}}\lVert {\mathbf{M}}{-}\hat{\mathbf{M}}\rVert_{2}$, 
$\mathcal{L}^{\text{J}}_\text{3D}{=}\tfrac{1}{N_J}\lVert {\mathbf{J}}{-}\hat{\mathbf{J}}\rVert_{2}$,
%
%
where $N_{M}$, $N_{J}$ is the number of 
mesh vertices and landmarks, respectively. 
%

%

\begin{table}[t!]
  \centering
  \begin{tabular}{c@{}c} 
  \raisebox{-.55\height}{\includegraphics[width=0.52\linewidth]{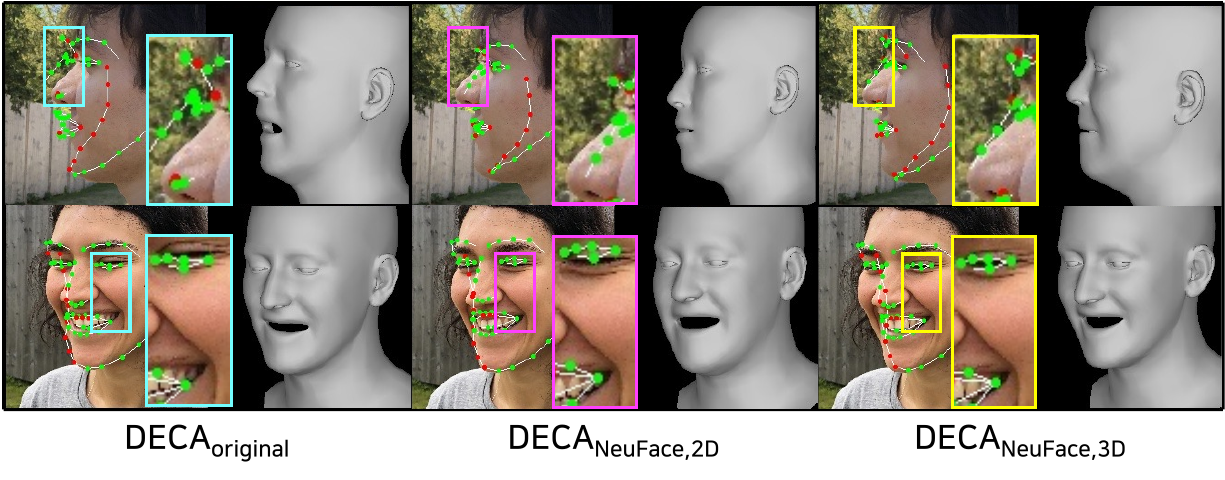}} & \hspace{1.5mm}
  \resizebox{0.45\linewidth}{!}{%
        \begin{tabular}{l cccc} 
        \toprule
        & & \multicolumn{3}{c}{Error [\emph{mm}] ($\downarrow$)}\\
        \cmidrule{3-5}
        Model & Test-opt & Median & Mean & Std \\
        \cmidrule{1-5}
        3DMM-CNN~\citep{tuan2017regressing}\textsubscript{\,CVPR 2017} &  & 1.84 & 2.33 & 2.05 \\
        PRNet~\citep{feng2018joint}\textsubscript{\,ECCV 2018} &  & 1.50 & 1.98 & 1.88 \\
        RingNet~\citep{sanyal2019learning}\textsubscript{\,CVPR 2019} &  & 1.21 & 1.54 & 1.31 \\
        MGCNet~\citep{shang2020self}\textsubscript{\,ECCV 2020} &  & 1.31 & 1.87 & 2.63 \\
        3DDFA-V2~\citep{guo2020towards}\textsubscript{\,ECCV 2020} & \checkmark & 1.23 & 1.57 & 1.39 \\
        DenseLandmarks~\citep{wood2022dense}\textsubscript{\,ECCV 2022} & \checkmark & 1.02 & 1.28 & 1.08 \\
        MICA~\citep{MICA:ECCV2022}\textsubscript{\,ECCV 2022} & \checkmark & 0.90 & 1.11 & 0.92 \\
        \cmidrule{1-5}
        DECA\textsubscript{original}~\citep{DECA:Siggraph2021}\textsubscript{\,SIGGRAPH 2021} &  & 1.18 & 1.46 & 1.25 \\
        \rowcolor{Gray}
        DECA\textsubscript{NeuFace,2D} (Ours) &  & 1.15 & 1.44 & 1.26 \\
        \rowcolor{Gray}
        DECA\textsubscript{NeuFace,3D} (Ours) &  & \textbf{1.11} & \textbf{1.38} & \textbf{1.19} \\
        \bottomrule
        \end{tabular}} \\ 
        \small (a) & \small (b)
    \end{tabular}\vspace{-1.5mm}
  \caption{
  \textbf{Improving the face reconstruction accuracy.} 
  (a) NeuFace-dataset helps the model reconstruct more occlusion robust and expressive 3D faces than the original model. Green and red dots denote visible and invisible 3D landmarks, respectively.
  (b) As a result, DECA\textsubscript{NeuFace, 2D}, DECA\textsubscript{NeuFace,3D}
  achieve better 3D reconstruction accuracy than DECA\textsubscript{original}.
  }\vspace{-4.5mm}
  \label{tab:finetune}
\end{table}

\paragraph{Enhancement on 3D reconstruction accuracy}
By fine-tuning DECA~\citep{DECA:Siggraph2021} 
using the images of MEAD~\citep{kaisiyuan2020mead}, VoxCeleb2~\citep{voxceleb2} and CelebV-HQ~\citep{zhu2022celebvhq},
with and without
our 3D supervision,
we 
obtain
DECA\textsubscript{NeuFace,3D} and DECA\textsubscript{NeuFace,2D}.
%
%
Following the 
evaluation protocol of the NoW benchmark~\citep{RingNet:CVPR:2019},
we reconstruct 3D faces for the provided images via each model and report the 
3D 
reconstruction errors.
%
In \Tref{tab:finetune},
our DECA\textsubscript{NeuFace,3D} 
shows lower 3D reconstruction error than 
DECA\textsubscript{original}
and DECA\textsubscript{NeuFace,2D}.
%
%

\subsection{Learning 3D human facial motion prior}
\label{sec:humor}

%

%
A facial motion prior is a versatile tool to understand how human faces move over time. 
It can 
generate realistic motions or regularize temporal 3D reconstruction~\citep{rempe2021humor}.
Unfortunately, the lack of large-scale 3D face video datasets makes learning facial motion prior infeasible.
We tackle this by exploiting the scale, diversity, and naturalness of the 3D facial motions in our dataset.

\paragraph{Learning facial motion prior}
%
We learn a 3D facial motion prior using HuMoR~\citep{rempe2021humor} with simple modifications.
%
HuMoR is 
a conditional VAE~\citep{Sohn15CVAE} 
that learns the transition distribution of human 
body motion.
%
%
We represent the state of a 
facial motion sequence
as the combination of FLAME parameters and landmarks in the NeuFace-dataset
and train the dedicated face motion prior, called HuMoR-Face.
%
%
We train three motion prior models (HuMoR-Face) with 
different training datasets, \ie, VOCASET~\citep{VOCA2019}, 
NeuFace\textsubscript{MEAD}, and NeuFace\textsubscript{VoxCeleb2}.
Please refer to supplementary material and HuMoR~\citep{rempe2021humor} for the details.

\begin{wrapfigure}{r}{0.6\linewidth}
    \centering
    \vspace{-3mm}
    \includegraphics[width=\linewidth]{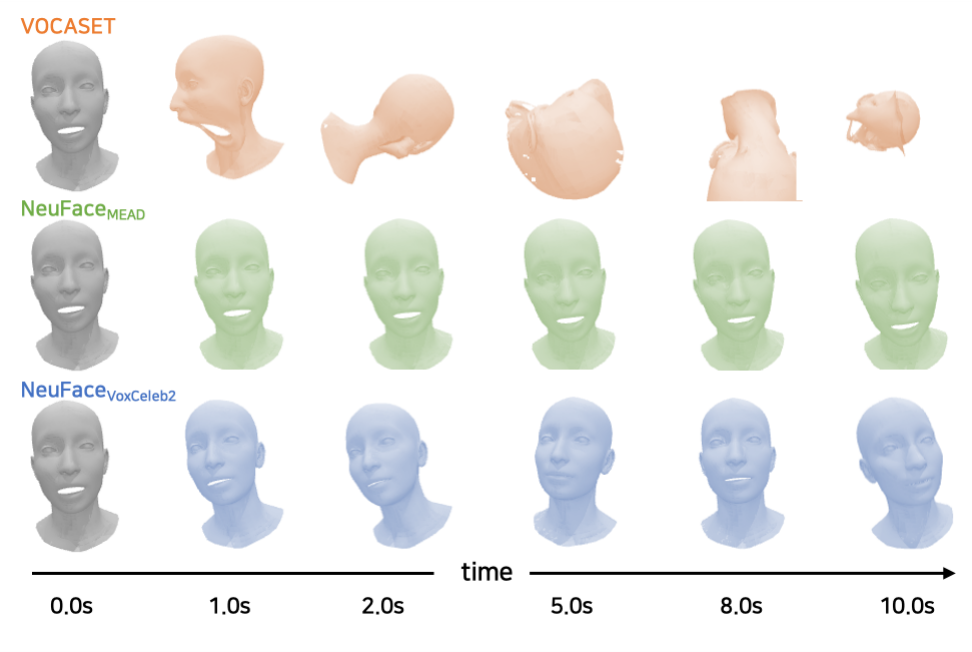}\vspace{-3mm}
    \caption{\textbf{Long-term facial motion generation using learned 
    motion prior.}
    %
    %
    The motion prior trained with small-scale, diversity-limited \orange{VOCASET} fails to generate natural motion, while the motion prior trained with \blueyw{NeuFace\textsubscript{VoxCeleb2}} generates diverse and natural long-term facial motion.}
    \label{fig:humor}
\end{wrapfigure}

\paragraph{Long-term face motion generation}
%
%
%
We evaluate the validity and generative power of the learned motion prior by
generating long-term 3D face motion sequences ($10.0s$).
%
Long-term motions are generated by auto-regressive sampling from the learned prior, given only a starting frame as the condition (see \Fref{fig:humor}).
%
%
VOCASET provides small-scale,
in-the-lab captured meshes, thus limited in motion naturalness and facial diversity.
%
Accordingly, the HuMoR-Face trained with VOCASET 
fails to learn a valid human facial motion prior and generates unnatural motion.
Using only the subset, NeuFace\textsubscript{MEAD}, 
the long-term stability of head motion has significantly enhanced. 
We attribute such high quality prior to 
the benefit
of the NeuFace-dataset: 
\emph{large-scale} facial motion annotations.
Further, exploiting diverse \emph{in-the-wild}, \emph{dynamic}, and \emph{natural} motion annotation from NeuFace\textsubscript{VoxCeleb2} 
helps HuMoR-Face learn real-world motion prior
and surprisingly generate much diverse and dynamic motions.
\section{Conclusion}
We develop NeuFace, an optimization for generating accurate and spatio-temporally consistent 3D face
mesh pseudo-labels on videos with provable optimal guarantee.
Moreover, with the technique, we build the NeuFace-dataset, a large-scale 3D face meshes paired with in-the-wild 2D videos.
%
%
We demonstrate the potential of 
the
diversity and naturalness of our NeuFace-dataset as a training dataset
to learn generative 3D facial motion prior. 
%
Also, we improve the reconstruction accuracy of a \emph{de-facto} standard 3D face reconstruction model using our dataset.
We expect NeuFace to open up new opportunities by providing large-scale, real-world 3D face video data, the NeuFace-dataset, as a reliable data curation method.

%

%
%

\subsection*{Acknowledgment}
This work was supported by Institute of Information \& communications Technology Planning \& Evaluation (IITP) grant funded by the Korea government(MSIT) (No.RS-2023-00225630, Development of Artificial Intelligence for Text-based 3D Movie Generation; and No.2022-0-00290, Visual Intelligence for Space-Time Understanding and Generation based on Multi-layered Visual Common Sense).

\subsection*{Ethics Statement}
\label{sec:supp_discussion}

%
For face reconstruction tasks and datasets, the diversity of race or ethnicity, gender, appearance, and actions is an important 
topic to discuss~\citep{Wang_2019_ICCV, zhu2022celebvhq}. 
%
Existing 3D face video datasets~\citep{MICA:ECCV2022, COMA:ECCV18, VOCA2019} typically have limited diversity regarding
ethnicity, gender, appearance, and actions. 
Such 3D face datasets rarely provide video pairs, but with 
artificial facial markers attached to human faces 
and a small set of identities.
On the other hand, our NeuFace-dataset mitigates such issues since our dataset is 
acquired on top of 
large-scale in-the-wild face video datasets, 
which typically 
rely on internet videos.
Such video datasets are 
diverse in terms of 
ethnicity, gender, facial appearances, and actions when compared to 
the small/medium-scale 3D facial motion capture datasets.
Since our dataset is acquired based on the existing public video datasets~\citep{kaisiyuan2020mead,voxceleb2,zhu2022celebvhq}, all the rights, licenses, and permissions follow the original datasets.
%
%
Moreover, we will release the NeuFace-dataset by providing the reconstructed 3DMM parameters without the actual facial video frames. 
NeuFace-dataset does not contain identity-specific metadata and facial texture maps. 
Nonetheless,  per-identity shape coefficients can give a rough guide about human facial shape. 
Thus, we will release our dataset for research purposes only.

\bibliography{iclr2024_conference}
\bibliographystyle{iclr2024_conference}

\clearpage
\newtheorem{assume}{Assumption}
\newtheorem{definition}{Definition}
\newtheorem{lemma}{Lemma}

\setcounter{section}{0}
\setcounter{figure}{0}
\setcounter{table}{0}
\setcounter{equation}{0}
\setcounter{theorem}{0}
\setcounter{proposition}{0}

\renewcommand\thesection{\Alph{section}}
\renewcommand\thefigure{S\arabic{figure}}
\renewcommand{\thetable}{S\arabic{table}}
\renewcommand\theequation{\alph{equation}}
\renewcommand{\colorref}[1]{{\color{black}{#1}}}

\appendix

\section*{Appendix}
We present additional analysis, results, and experiments that are not
included in the main paper due to the space limit.
Also, the attached video explains and demonstrates the main idea of NeuFace and video samples for the NeuFace-dataset.

\section{Analysis on NeuFace}
\label{sec:supp_analysis}
In this section, we introduce and validate our design choices for NeuFace optimization, through analysis.
Specifically, we build a strong baseline 
and support our choice of ``re-parameterized'' face mesh optimization method 
for NeuFace in \Sref{sec:supp_flame_vs_ours}.
Next, we provide a proof sketch of the provable global minima convergence of NeuFace optimization in \Sref{sec:supp_convergence}.
Also, we analyze and discuss the effect of each loss function in \Sref{sec:supp_ablation}.

\subsection{FLAME fitting vs. NeuFace optimization}
\label{sec:supp_flame_vs_ours}
Recall that NeuFace re-parameterizes the 3DMM, \ie, FLAME~\citep{FLAME:SiggraphAsia2017} to the neural parameters (represented as DECA~\citep{DECA:Siggraph2021}), then optimizes over them to obtain accurate 3D face meshes for videos.
Following the prior arts in the parametric human body 
reconstruction literature~\citep{bogo2016smplify, SMPL-X:2019, kolotouros2019spin}, there exists a simple method to optimize the parametric model; 3DMM parameter fitting.
Thus, we implement FLAME fitting as a solid baseline and compare the quantitative and qualitative results with NeuFace optimization to analyze and support our choice of neural re-parameterization.

\paragraph{Details of baseline  FLAME fitting}
%
%
%
Given the initial FLAME and camera parameters,
$[\boldsymbol{\Theta_{\bb}}, \bp_{\bb}] = [\br_{\bb}, \btheta_{\bb}, \bbeta_{\bb}, \bpsi_{\bb}, \bp_{\bb}]$,
we implement the direct FLAME optimization as:
\begin{equation*}
    \label{eq:supp_baseline}
    [\boldsymbol{\Theta_{\bb}^{*}}, \mathbf{p}_{\bb}^{*}] = 
    \argmin_{\boldsymbol{\Theta_{\bb}, \mathbf{p}_{\bb}}} \,\,
    \mathcal{L}_{\textrm{2D}} + \lambda_{\textrm{temp}}\mathcal{L}_{\textrm{temporal}} + \lambda_{\textrm{view}}\mathcal{L}_{\textrm{multiview}}
    + \lambda_{\mathbf{r}}\mathcal{L}_{\mathbf{r}}
    + \lambda_{\boldsymbol{\theta}}\mathcal{L}_{\boldsymbol{\theta}}
    + \lambda_{\boldsymbol{\beta}}\mathcal{L}_{\boldsymbol{\beta}}
    + \lambda_{\boldsymbol{\psi}}\mathcal{L}_{\boldsymbol{\psi}},
\end{equation*}
%
where the losses $\mathcal{L}_{\textrm{2D}}$, $\mathcal{L}_{\textrm{temporal}}$, and $\mathcal{L}_{\textrm{multiview}}$ are identical to the 
losses discussed in the main paper (Eqs.~\ref{eq:l_2d},\ref{eq:l_temp},\ref{eq:l_mv}).
We can obtain the initial FLAME parameters 
for the optimization
in two ways: (1) initialize from mean parameters and (2) initialize from pre-trained DECA~\citep{DECA:Siggraph2021} predictions. 
%
We empirically found that initialization with mean FLAME parameters frequently fails
when the input images
contain extreme head poses. 
Thus, we choose to initialize FLAME parameters from the 
pre-trained DECA predictions, thus providing 
a
plausible 
initialization for a fair comparison.
%
Also, following the convention~\citep{FLAME:SiggraphAsia2017}, 
we optimize FLAME parameters in a coarse-to-fine manner.
For the earlier stage, we fix FLAME parameters that control local details, \ie, $\boldsymbol{\theta}_{\bb}$, $\boldsymbol{\beta}_{\bb}$, and $\boldsymbol{\psi}_{\bb}$, and optimize the global head orientation, $\mathbf{r}_{\bb}$, and camera parameters $\mathbf{p}_{\bb}$.
Then we fix camera parameters and optimize other FLAME parameters 
jointly at a later stage to fit the local details.
%


%

%
Since we initialize FLAME parameters and camera parameters from the pre-trained DECA predictions, \ie, initial $[\boldsymbol{\Theta_{\bb}}, \mathbf{p}_{\bb}]$ in \Eref{eq:baseline}.
Accordingly, the meshes obtained by the FLAME fitting achieve better spatio-temporal consistency and 2D landmark accuracy 
than the meshes obtained by a pre-trained DECA without any post-processing.
%
%
%
%

\begin{figure}[ht!]
  \centering
  \begin{tabular}{c@{}c} 
  {\includegraphics[width=0.48\linewidth]{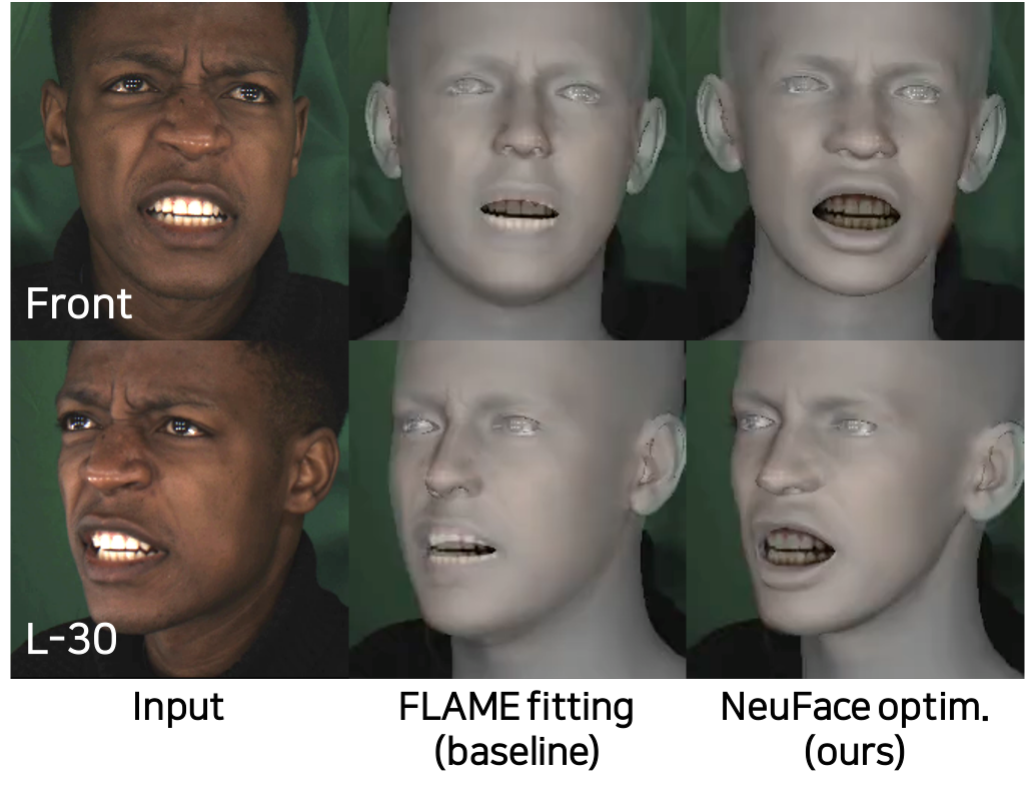}} & \hspace{1.5mm}
  {\includegraphics[width=0.48\linewidth]{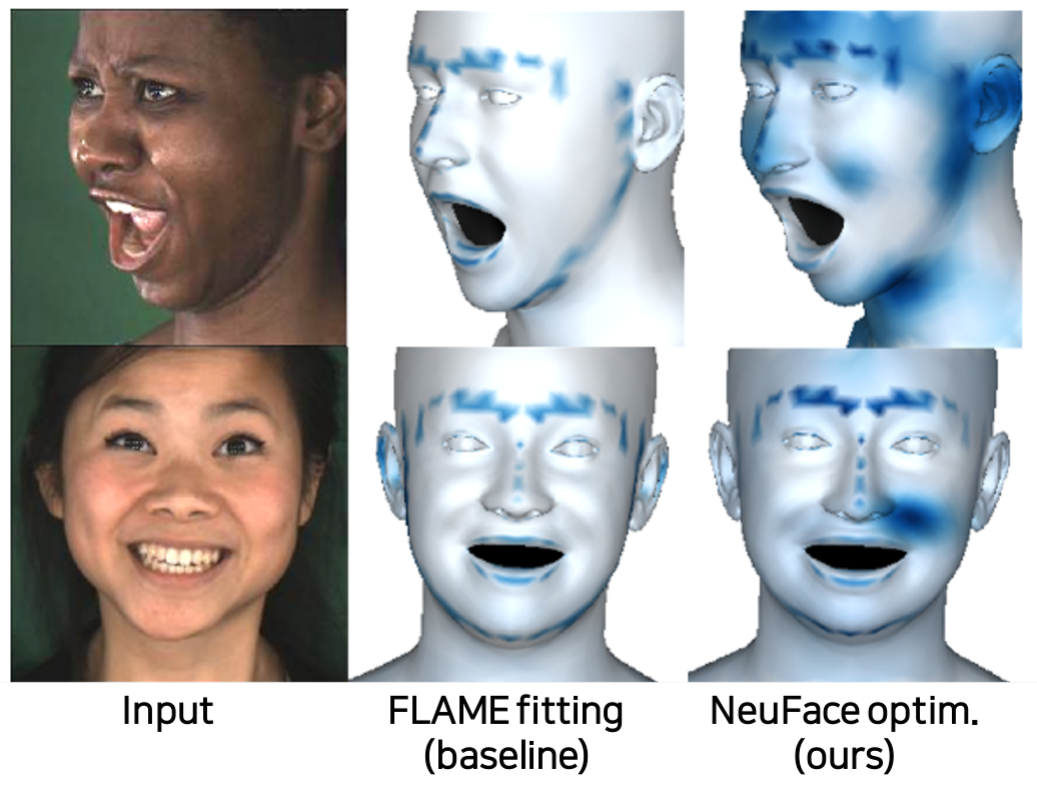}}\vspace{-1.5mm}\\
        \small (a) & \small (b)
    \end{tabular}\vspace{-3mm}
  \caption{
  \textbf{FLAME fitting vs. NeuFace optimization.} 
  (a) NeuFace optimization obtains much expressive and pixel-level aligned meshes than the baseline FLAME fitting.
  (b) We observe NeuFace optimization induces richer and data-dependent gradients compared to the sparse gradients of baseline.
  }
  \label{fig:supp_baseline}
\end{figure}


\paragraph{Qualitative result}
In \Fref{fig:supp_baseline}\colorref{(a)}, 
NeuFace-dataset contains much expressive and image-aligned meshes, \eg, wrinkles and face boundaries. 
On the other hand, the meshes obtained by the direct FLAME optimization show mean shape-biased 3D faces. 
%
%
We explain such results in terms of the 
data-independency of the direct FLAME optimization.
%
The baseline method requires several regularization terms based on the prior, pre-built from external 3D datasets, which is data-independent (\Eref{eq:baseline}).
Such data-independent regularization 
encourages the optimized FLAME parameters to stay close to the 
mean of 
parameter distributions, regardless of the 
facial characteristics on input images.
Balancing such regularization terms with other losses is cumbersome and prone to obtain mean shape faces.
%
%
%


In contrast, recall that NeuFace re-parameterizes the FLAME parameters as the pre-trained neural network, such as DECA~\citep{DECA:Siggraph2021}. 
Such re-parameterization allows NeuFace to 
update face meshes in an input-image-conditioned manner,
called data-dependent mesh update (\Sref{sec:why_reparam} in the main paper).
Figure~\ref{fig:supp_baseline}\colorref{(b)} visualizes the data-dependent gradient of our optimization. 
While the baseline shows similar gradient patterns throughout three input images, NeuFace optimization produces diverse gradient maps according to input images. Such rich and data-dependent gradients consider high-dimensional visual features induced from the input RGB values in the optimization process, yielding accurate and expressive 3D faces.
%
%
%
%

%
%

\subsection{Proposition 1: Convergence Property to Global Minima}
\label{sec:supp_convergence}
Proposition 1 in this work is a straightforward variation of the main result of Allen-Zhu~\etal\citep{allenzhu2019convergence}.
Before providing the proof sketch of Proposition 1. We describe the assumptions needed to prove Proposition 1.
For simplicity, we consider a simple $l_2$-based regression loss and a $L$-layer fully connected ReLU network $\Phi_{\overrightarrow{\mathbf{W}}}(\cdot)$ having a uniform weight width size of $m$.
For the training data consisting of vector pairs $\{(x_i, y_i^*)\}_{i\in[n]}$, the network has the batched input of $\bx{=}\{x_i\}_{i\in[n]}$, where $n$ is the batch size, $x_i {\in} [0,1]^{d_{in}}$, and $y_i^*\in \Real^{d_{out}}$, the output of $\Phi_{\overrightarrow{\mathbf{W}}}({x}) \in \Real^{d_{out}}$, and the weights $\overrightarrow{\bW}=(\bW_1 {\in} \Real^{m \times d_{in}},\bW_{\{2,\cdots, L-1\}} {\in} \Real^{m \times m},\bW_{L} {\in} \Real^{d_{out} \times m})$.\footnote{We can consider that the biases are included in $\{\bW\}$ without loss of generality.}

\begin{assume}\label{thm:assume1}
Without loss of generality, $\forall i, \|x_i\|=1$ and $\|y_i^*\| \leq O(1)$.
\end{assume}
\begin{assume}\label{thm:assume2}
The pretrained neural network weights $\overrightarrow{\bW}^{(0)}$ are assumed to be started from the values distributed normally, \ie, considered as a sample instance from a Gaussian distribution. Specifically, $[\bW_{l}^{(0)}]_{i,j} \sim \mathcal{N}(0,2/\mathtt{row}[\bW_{l}])$ for $l\in \{1, \cdots, L-1\}$ and $[\bW_{L}^{(0)}]_{i,j} \sim \mathcal{N}(0,1/\mathtt{row}[\bW_{L}])$ for every $(i,j)$, where the operator $\mathtt{row}[\cdot]$ returns the row size of the input matrix.
\end{assume}

Assumption~\ref{thm:assume2} appears to be restrictive by those standard deviations, but it is not.
The assumptions cover a fairly broad range of weight distribution scenarios.
For larger standard deviations, we can always set a small norm for $x$'s in Assumption~\ref{thm:assume1} without loss of generality, and vice versa.

Under these assumptions, we restate Proposition 1 in the main paper. 
\begin{proposition}[Global Convergence]
For any $\epsilon \in (0,1],\,\,\delta {\in} \left( 0, O(\tfrac{1}{L})\right]$, given an input data $\{\mathbf{x}, \by\}$ and 
the neural network $\Phi_{\overrightarrow{\mathbf{W}}}(\cdot)$ over-parameterized such that $m\geq \Omega\left(\mathtt{poly}(n, L,\delta^{-1})\cdot d_{out}\right)$, 
consider optimizing the non-convex regression problem: $\argmin_{\overrightarrow{\mathbf{W}}}\calL({\overrightarrow{\mathbf{W}}}) = \tfrac{1}{2}\|\Phi_{\overrightarrow{\mathbf{W}}}(\mathbf{x}) - \mathbf{y}\|_2^2$.\\
Under the above assumptions, with high probability, the gradient descent algorithm with the learning rate $\rho = \Theta\left( \tfrac{d\delta}{\mathtt{poly}(n,L)m} \right)$ finds a point $\overrightarrow{\mathbf{W}}^*$ such that $\calL({\overrightarrow{\mathbf{W}}^*})\leq \epsilon$ in polynomial time.
\end{proposition}

\noindent
\emph{Proof sketch.}
We first introduce the following useful lemmas needed to prove the proposition.

\begin{lemma}[Theorem 3 in~\citep{allenzhu2019convergence}]\label{thm:nocriticalpt}
With probability at least $1-e^{-\Omega(m/ \Omega(\mathtt{poly}(n,L,\delta^{-1}))}$,
it satisfies for every $\ell \in [L]$, every $i\in[n]$, and every $\overrightarrow{\bW}$ with $\lVert\overrightarrow{\bW}-\overrightarrow{\bW}^{(0)}\rVert_{2}\le\tfrac{1}{\mathtt{poly}(n,L,\delta^{-1})}$, where  $\|\overrightarrow{\bW}\|_{2} = \max_{l\in [L]} \|\bW_l\|_{2}$, 
\begin{align*}
    \Omega\left(\calL(\overrightarrow{\bW})\cdot\frac{\delta m}{dn^2}\right) \leq 
    \lVert \nabla \calL(\overrightarrow{\bW}) \rVert^{2}_{F}\le O\left(\calL(\overrightarrow{\bW})\cdot\frac{Lnm}{d}\right).
\end{align*}
\end{lemma}

This lemma suggests that, when we are close to the starting point $\overrightarrow{\bW}^{(0)}$ of the neural network, 
there is no saddle point or critical point of any order.
Specifically, for example, given a fixed $\delta, d, n$ and $L$, when we have the same error $\calL$ for two different neural networks with respective widths of $m_1$ and $m_2$, where $m_1 < m_2$, then
the lower bound of the gradient with $m_2$ is larger than that of $m_1$ with a better chance. 
This means that neural networks with larger widths are likely to have a lower chance of local minima.

This hints that any local search (\eg, gradient descent) does not suffer from any local minima or saddle points for larger $m$, which implies a more likely chance of avoiding local minima, \ie, finding global minima. 
However, the local search does not guarantee to decrease the loss function yet.

With the favorable property of Lemma~\ref{thm:nocriticalpt}, if we have an additional guarantee of loss decrease with gradient descent, we can prove the convergence to global minima.
To derive objective-decrease guarantee in optimization theory, a notion of smoothness is typically needed; thus, we introduce the following lemma.

\begin{lemma}[Theorem 4 of~\citep{allenzhu2019convergence}]\label{thm:semismooth}
With probability at least $1-e^{-\Omega(m/\mathtt{poly}(L,\log m))}$, we have:
for every $\overrightarrow{\bW}^{\dagger}\in(\mathbb{R}^{m \times m})^{L}$ with 
$\lVert\overrightarrow{\bW}^{\dagger}-\overrightarrow{\bW}^{(0)}\rVert_{2}\le\frac{1}{\mathtt{poly}(L,\log m)}$,\\
and for every $\overrightarrow{\bW}^{\prime}\in(\mathbb{R}^{m \times m})^{L}$ with 
$    \lVert\overrightarrow{\bW}^{\prime}\rVert_{2}\le\frac{1}{\mathtt{poly}(L,\log m)},$
the following inequality holds
\begin{align*}
\calL(\overrightarrow{\bW}^{\dagger}{+}\overrightarrow{\bW}^{\prime})\le&\,\, \calL(\overrightarrow{\bW}^{\dagger})+\left<\nabla \calL(\overrightarrow{\bW}^{\dagger}), \overrightarrow{\bW}^{\prime}\right> + O(\tfrac{nL^{2}m}{d})\lVert\overrightarrow{\bW}^{\prime}\rVert^{2}_{2}
+ \tfrac{\mathtt{poly}(L)\sqrt{nm\log m}}{\sqrt{d}}\cdot\lVert\overrightarrow{\bW}^{\prime}\rVert_{2}
\sqrt{\calL(\overrightarrow{\bW}^{\dagger})}.
\end{align*}
\end{lemma}

This lemma states the semi-smoothness property of the objective function $\calL$ \wrt $\Phi_{\overrightarrow{\mathbf{W}}}(\cdot)$ to take into account 
non-smoothness introduced by ReLU activation in $\Phi_{\overrightarrow{\mathbf{W}}}(\cdot)$.
The semi-smoothness looks similar to the Lipschitz smoothness except for the first order term $\lVert\overrightarrow{\bW}^{\prime}\rVert_{2}$. Interestingly, when we increase $m$, the increasing rate of the first order term is much slower than that of the second order term; thus, the second order term becomes dominant compared to the first order one, and the semi-smoothness approaches closer to the Lipschitz smoothness. 
This means that the neural network is smoother as $m$ goes larger.

Under the assumption that $\|\overrightarrow{\bW}^{(t)}-\overrightarrow{\bW}^{(0)}\|_F$ is small (will be verified later), the next step is to combine Lemma~\ref{thm:semismooth} with gradient descent to derive the loss-decrease guarantee. 
Denoting $\nabla_t = \nabla\calL(\overrightarrow{\bW}^{(t)})$, the gradient descent update rule is defined as: $\overrightarrow{\bW}^{(t+1)} = \overrightarrow{\bW}^{(t)} - \rho\nabla_t$ for a learning rate $\rho>0$.
Then, from Lemma~\ref{thm:semismooth}, 
putting $\overrightarrow{\bW}^{(t+1)}=\overrightarrow{\bW}^{\dagger}+\overrightarrow{\bW}^{\prime}$ and $\overrightarrow{\bW}^{(t)}=\overrightarrow{\bW}^{\dagger}$, \ie, $\overrightarrow{\bW}^{\prime}=-\rho\nabla_t$,
we have 
\begin{flalign}
\calL(\overrightarrow{\bW}^{(t+1)})\le \calL(\overrightarrow{\bW}^{(t)}) - \rho\|\nabla_t\|_F^2 + \rho^2 C_1\|\nabla_t\|^{2}_{2}+ {\rho}C_2\|\nabla_t\|_{2}
\sqrt{\calL(\overrightarrow{\bW}^{(t)}))}\nonumber&&
\end{flalign}
\hfill (where $C_1 = O(\tfrac{nL^{2}m}{d})$, $C_2 = \tfrac{\mathtt{poly}(L)\sqrt{ nm\log m}}{\sqrt{d}}$)
\begin{flalign}
\phantom{\calL(\overrightarrow{\bW}^{(t+1)})} \le \,\, \calL(\overrightarrow{\bW}^{(t)}) - \rho\|\nabla_t\|_F^2 + \left( \rho^2 C_1 O\left(  \tfrac{nm}{d} \right) +
\rho C_2
\sqrt{  O\left( \tfrac{nm}{d} \right)} \right) \calL(\overrightarrow{\bW}^{(t)}) \nonumber&&
\end{flalign}
\hfill ($\|\nabla_t\|_2^2 \leq \max\limits_{l\in[L]} \| \nabla_{\bW_l} \calL(\overrightarrow{\bW}^{(t)}))\|_F^2 \leq  O\left(  \tfrac{nm}{d} \right) \calL(\overrightarrow{\bW}^{(t)})$ from the  upper bound in Lemma~\ref{thm:nocriticalpt})
\begin{flalign}
\phantom{\calL(\overrightarrow{\bW}^{(t+1)})} = \,\, (1 + C_3)  \calL(\overrightarrow{\bW}^{(t)}) - \rho\|\nabla_t\|_F^2 &&
\textrm{\hfill(where $C_3 = \rho^2 C_1 O\left(  \tfrac{nm}{d} \right) +
\rho C_2
\sqrt{  O\left( \tfrac{nm}{d} \right)}$)}
\nonumber
\end{flalign}
\begin{flalign}
\phantom{\calL(\overrightarrow{\bW}^{(t+1)})} \leq \,\, \left( 1 
- 
\Omega\left( \frac{ \rho \delta m}{dn^2}\right)  \right) \calL(\overrightarrow{\bW}^{(t)})\nonumber&&
\end{flalign}
\hfill (by the gradient lower bound from Lemma~\ref{thm:nocriticalpt} and our choice of $\rho$, \eg, $\rho=\Theta(\tfrac{d\delta}{n^4 L^2 m})$)

\vspace{2mm}
When we choose the parameters such that $ \Omega\left( \frac{\rho \delta m}{dn^2}\right)   {\in} (0,1)$, we have $\calL(\overrightarrow{\bW}^{(t+1)}) {<} \calL(\overrightarrow{\bW}^{(t)})$. 
In other words, there exists $T>0$ such that $\calL({\overrightarrow{\mathbf{W}}^{(T)}})\leq \epsilon$.
Examples of convenient parameter choices of $m, \rho$, and $T$ in polynomial orders are suggested in \citet{allenzhu2019convergence} to hold $ \Omega\left( \frac{\rho \delta m}{dn^2}\right)     \in (0,1)$ and the small value of $\|\overrightarrow{\bW}^{(t)}{-}\overrightarrow{\bW}^{(0)}\|_F$ for every $t$.
This concludes the proof sketch of finding a point ${\overrightarrow{\mathbf{W}}^{*}} {=} {\overrightarrow{\mathbf{W}}^{(T)}}$ such that $\calL({\overrightarrow{\mathbf{W}}^*}) \leq \epsilon$. \hfill $\square$\vspace{1mm}

\paragraph{Remark 1: Global optimality}
In Proposition 1, we can set $\epsilon$ arbitrarily small. With a very small $\epsilon$, it suggests that a converged point $\overrightarrow{\mathbf{W}}^*$ is a global minimum.

\paragraph{Remark 2: The radius condition between the initial weights and updated one}
The spectral radius bounds for $\|\overrightarrow{\bW}-\overrightarrow{\bW}^{(0)}\|_F$ required in 
Lemmas~\ref{thm:semismooth} and \ref{thm:nocriticalpt} appear to be small, it is sufficiently large enough to completely change the output of the model, considering the large width of size $m$ and the standard deviation $\tfrac{1}{\sqrt{m}}$ of weight entries in Assumption~\ref{thm:assume2}.

\paragraph{Remark 3: Other architectures}
\citet{allenzhu2019convergence,du2019gradient} present the recipes to convert the $L$-layer fully connected networks to convolutional neural networks and to ResNet by sacrificing the complexity of proof. Thus, the conclusion of the provable guarantee does not change with such architectural changes.
Thus, the architectures we experimented provably comply with the conclusion of Proposition 1 up to the choice of the parameters, \ie, global convergence.

\paragraph{Remark 4: Other losses}
In the above proof sketch, one of the important pieces is the semi-smoothness in Lemma~\ref{thm:semismooth}.
While we discuss only with the simple $l_2$ regression loss function, 
fortunately, the semi-smoothness already encompasses any choice of Lipschitz smooth cases for the loss functions.
Thus, as long as the choice of the loss function is Lipschitz smooth, the replacement of the loss function does not alter the conclusion of Proposition 1 even for non-convex losses except the choice of parameters. 
This hints that our choice of the multi-task loss in \Eref{eq:obj} provably complies with the conclusion of Proposition 1 except the choice of the parameters.

\paragraph{Remark 5: $L$ vs.~$m$ for the over-parameterization}
For designing the over-parameterized architecture, one can control two different parameters $L$ and $m$.
Obviously, the high probability is achieved with larger $m$ rather than $L$, but more importantly, the local minima smoothing phenomenon suggested in the lower bound of Lemma~\ref{thm:nocriticalpt} is independent to $L$.

\subsection{Ablation on loss functions}
\label{sec:supp_ablation}
We conduct ablation studies to analyze the effect of our proposed spatio-temporal consistency losses in the {NeuFace} optimization, $\mathcal{L}_{\text{multiview}}$, and $\mathcal{L}_{\text{temporal}}$. 
We evaluate the quality of the meshes obtained by optimizing each of the loss configurations.
All the experiments are conducted on the same validation set as the \Tref{evaluation_results} of the main paper.

\begin{table*}[h]
  \centering
  \resizebox{1\linewidth}{!}{
  \begin{tabular}{ccc cccc ccc ccc}
  \toprule
\multicolumn{3}{c}{NeuFace} &\multicolumn{4}{c}{MEAD~\citep{kaisiyuan2020mead}} & \multicolumn{3}{c}{VoxCeleb2\citep{voxceleb2}} & \multicolumn{3}{c}{CelebV-HQ~\citep{zhu2022celebvhq}} \\
     \cmidrule(lr){1-3} \cmidrule(lr){4-7} \cmidrule(lr){8-10} \cmidrule(lr){11-13}
    \multicolumn{1}{l}{$\mathcal{L}_{\text{2D}}$}&\multicolumn{1}{l}{$\mathcal{L}_{\text{multiview}}$}&\multicolumn{1}{l}{$\mathcal{L}_{\text{temporal}}$} & \multicolumn{1}{c}{CVD $\downarrow$} & \multicolumn{1}{c}{$\text{MSI}_\text{3D}^\text{L}\uparrow$} &\multicolumn{1}{c}{$\text{MSI}_\text{3D}^\text{V}\uparrow$} & \multicolumn{1}{c}{NME $\downarrow$} & \multicolumn{1}{c}{$\text{MSI}_\text{3D}^\text{L}\uparrow$} &\multicolumn{1}{c}{$\text{MSI}_\text{3D}^\text{V}\uparrow$} & \multicolumn{1}{c}{NME $\downarrow$} & \multicolumn{1}{c}{$\text{MSI}_\text{3D}^\text{L}\uparrow$} &\multicolumn{1}{c}{$\text{MSI}_\text{3D}^\text{V}\uparrow$} & \multicolumn{1}{c}{NME $\downarrow$} \\
  \cmidrule{1-13}
\checkmark&&  & 0.333 & 0.015&0.022 & \textbf{2.44} & 0.001 & 0.001 & 2.71 & 0.001 & 0.003 &3.88\\
\checkmark&\checkmark&  & 0.110 & 0.015&0.023 &2.49 & - & - & - & - & - & -\\
\checkmark&&\checkmark& 0.294 & 0.176&0.250 & 2.56 & \textbf{0.095} & \textbf{0.137} & \textbf{2.19} & \textbf{0.54} & \textbf{0.074} & \textbf{2.55}\\
\checkmark&\checkmark&\checkmark&\textbf{0.103} & \textbf{0.206} & \textbf{0.305} & 2.58 & - & -& - & -&-&-\\
\cmidrule{1-13}
\multicolumn{3}{c}{DECA~\citep{DECA:Siggraph2021}}  & 0.209 & 0.011&0.016 & 4.65 & 0.028 & 0.044 & 4.78 & 0.012 & 0.018 &5.34\\
    \bottomrule
  \end{tabular}}
  \caption{\textbf{Ablation results on the different loss functions.} 
  We evaluate the effect of our proposed spatio-temporally consistent losses by changing the configurations of the loss combinations. 
  Optimizing full loss functions shows favorable results on CVD and NME while outperforming MSI compared to other configurations. 
  We cannot analyze the effect of $\mathcal{L}_{\text{multiview}}$ for VoxCeleb2 and CelebV-HQ since they are taken in the single-camera setup. 
  $\emph{Abbr.}$ L: landmark, V: vertex.
  }
\label{tab:ablation}
\end{table*}

\begin{wrapfigure}{r}{0.45\linewidth}
    \centering
    \vspace{-4.5mm}
    \includegraphics[width=0.9\linewidth]{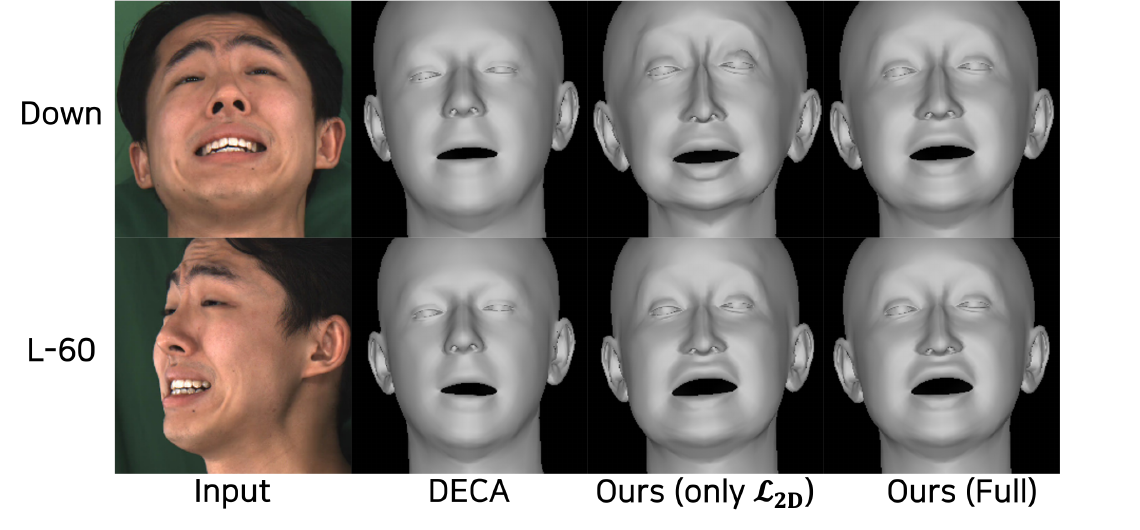}\vspace{-3mm}
    \caption{\textbf{Effect of the $\mathcal{L}_{\text{multiview}}$.} 
    }
    \vspace{-1mm}
    \label{fig:ablation}
\end{wrapfigure}

\paragraph{Only $\mathcal{L}_{\text{2D}}$}
We start by optimizing the loss function with only 2D facial landmark re-projection error, $\mathcal{L}_{\text{2D}}$.
As we optimize $\mathcal{L}_{\text{2D}}$, the obtained meshes achieve lower NME than DECA over the whole validation set (see \Tref{tab:ablation}).
However, we observe that this may break both multi-view and temporal consistencies, degrading the CVD and MSI compared to that of DECA.
The qualitative results in \Fref{fig:ablation} also show that optimizing only $\mathcal{L}_{\text{2D}}$ reconstructs more expressive but inconsistent meshes over different views.
Therefore, we propose loss functions that can induce multi-view and temporal consistencies 
to the meshes during the optimization.

\paragraph{$\mathcal{L}_{\text{2D}}{+}\mathcal{L}_{\text{multiview}}$}
By optimizing $\mathcal{L}_\text{multiview}$ along with $\mathcal{L}_{\text{2D}}$,
we observe that CVD gets significantly lower than DECA and the 
meshes
optimized with only $\mathcal{L}_{2D}$.
We have not optimized any regularization term which induces temporal consistency; thus, MSIs remain low. 
As discussed in the main paper, a trivial solution for achieving low CVD is to regress mean faces over different views.
However, 
the optimization
with $\mathcal{L}_{\text{2D}}{+}\mathcal{L}_{\text{multiview}}$ achieves low CVD while comparable NME to the 
$\mathcal{L}_{\text{2D}}$
optimization,
which proves not to be falling into a trivial solution.
Note that $\mathcal{L}_{\text{multiview}}$ can only be measured in the multi-camera setup, \eg MEAD.

\paragraph{$\mathcal{L}_{\text{2D}}{+}\mathcal{L}_{\text{temporal}}$}
As we optimize $\mathcal{L}_{\text{temporal}}$ with $\mathcal{L}_{\text{2D}}$, we observe substantial improvements in MSIs over the whole validation set.
Interestingly, jointly optimizing these two losses can further achieve better NME in VoxCeleb2 and CelebV-HQ datasets.
We postulate that, for in-the-wild challenging cases, \eg, images containing extreme head poses or diverse background scenes,
only 
optimizing
$\mathcal{L}_{\text{2D}}$ could fail to regress 
proper meshes, as it may generate meshes that break out of the facial regions.
%
On the other hand, 
$\mathcal{L}_{\text{temporal}}$
could prevent the regressed mesh from breaking out
from the facial regions to a certain extent. 

\begin{wrapfigure}{r}{0.6\linewidth}
    \centering
    \vspace{-4.5mm}
    \includegraphics[width=1\linewidth]{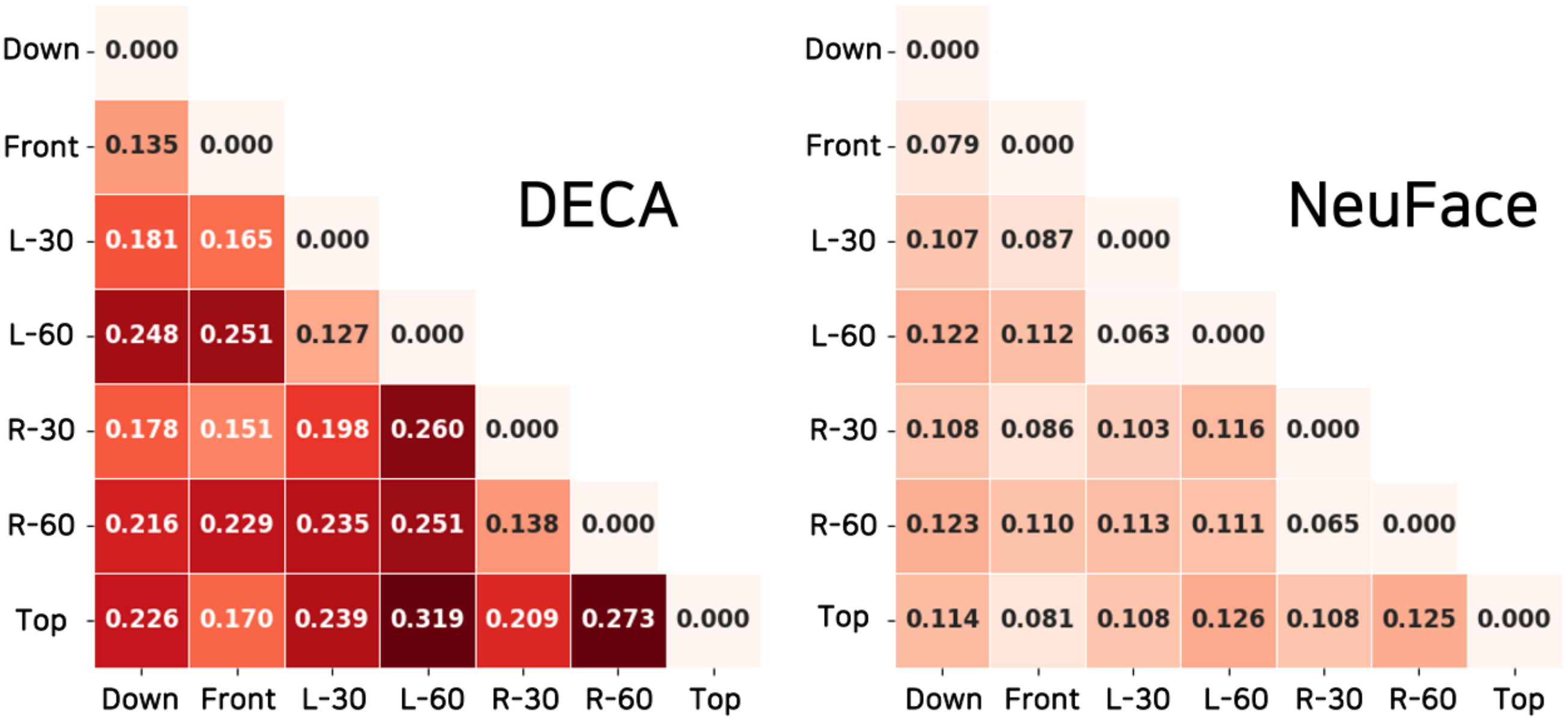}
    \caption{\textbf{Comparisons of cross-view vertex distance.} We quantitatively show the multi-view consistency of our method by averaging the cross-view vertex distance on the validation set of MEAD. L- and R- denote Left and Right, respectively, and 
    30 and 60 denote the view angles which the video is captured from.}
    \vspace{-3mm}
    \label{fig:quan}
\end{wrapfigure}

\paragraph{Full loss function}
With the observations of the effect on each proposed loss, we optimize the full loss functions (NeuFace optimization), $\mathcal{L}_{\text{2D}}{+}\mathcal{L}_{\text{spatial}}{+}\mathcal{L}_{\text{temporal}}$, on MEAD.
The quantitative results in \Tref{tab:ablation} show that NeuFace optimization
achieves comparable NME while outperforming CVD and MSIs compared to other settings.
As analyzed in the main paper, we postulate that our proposed spatio-temporal losses are mutually helpful for generating multi-view and temporally consistent meshes.

\begin{figure*}[h]
    \centering
    \includegraphics[width=\linewidth]{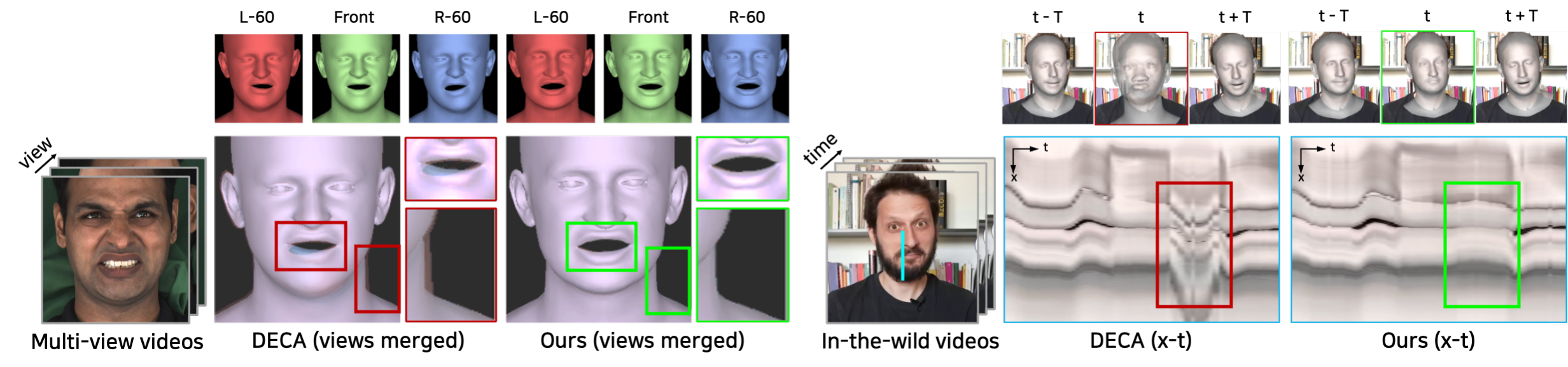}
    %
    %
    \caption{
    \textbf{Spatio-temporal consistency comparison.} 
    NeuFace reconstructs \emph{multi-view consistent 3D faces} (left); reduced misalignment for faces 
    (each view is 
    aligned to the canonical pose, color-coded
    and alpha-blended), 
    and \emph{temporally stabilized motion} (right); reduced jitter than the competing method 
    (we concatenate vertical cyan line in each frame along time).
    while being accurate.
    %
    %
    }
    \label{fig:teaser}
\end{figure*}

The advantage of jointly optimizing all the losses can also be found in the qualitative results; the reconstructed face meshes are well-fitted to its 2D face features, \eg, landmarks and wrinkles, and multi-view consistent (See \Fref{fig:ablation}).
In addition, we compare the view-wise averaged CVD between NeuFace optimization and DECA in \Fref{fig:quan}.
While DECA results in high CVD, especially for the views with self-occluded regions, such as Left-60 and Right-60, NeuFace shows significantly lower CVD on overall views.
%
We also evaluate
the meshes reconstructed by pre-trained DECA~\citep{DECA:Siggraph2021} for comparison (see \Fref{fig:teaser}).

\section{Implementation details}\label{sec:details}
We provide detailed configurations for 
implementations and experiments in the main paper.
%

\paragraph{NeuFace optimization details}
NeuFace optimization is composed of a
neural network $\Phi_{\textbf{w}}$ and the optimizing part.
For the network $\Phi_{\textbf{w}}$, we use a pre-trained DECA~\citep{DECA:Siggraph2021} or a pre-trained EMOCA~\citep{EMOCA:CVPR:2022} encoder network.
%
Overall optimization takes about 8 min. for $7$ views, ${\sim}120$ frames of videos, and about 2.5 min. for $1$ view, ${\sim}120$ frames of videos.

\paragraph{NeuFace-dataset acquisition} 
We provide reliable 3D face mesh annotations for large-scale face video datasets: MEAD~\citep{kaisiyuan2020mead}, VoxCeleb2~\citep{voxceleb2}, and CelebV-HQ~\citep{zhu2022celebvhq}.
We optimize our full objective (\Eref{eq:obj}) to acquire FLAME 
meshes for the datasets with a multi-view camera setup, \eg, MEAD.
Otherwise, we optimize (\Eref{eq:obj}) with $\lambda_{\text{view}}{=}0$.
We automatically discard the sequences if the optimization yields 
out-of-distribution shape parameters, \ie, the L2-norm of 
shape parameters deviates largely from the 
pre-built distribution~\citep{FLAME:SiggraphAsia2017}, $\lVert\boldsymbol{\beta}\rVert_2{>}1.0$, or 
if the 2D landmark detector~\citep{FAN} fails to capture the faces.
After subsequent human verification, 
the NeuFace-dataset
achieves ${<}0.1\%$ of failure rate
on upon criteria, supporting the reliable quality of our dataset.

\paragraph{Facial motion prior}
As one of our dataset's prominent applications, we introduced 
the learning of 3D facial motion prior, called HuMoR-Face (Sec.\colorref{5.1} in the main paper).
We first pre-process 3D face meshes in the NeuFace-dataset.
We compute root orientation, face pose angles, 3D landmark positions, and their velocities, respectively. 
Then, we represent the state of a moving human face as $\textbf{x}=[\boldsymbol{\phi}, \dot{\boldsymbol{\phi}},\boldsymbol{\theta}, \textbf{J}, \dot{\textbf{J}}]$, 
where $\boldsymbol{\phi}$, $\dot{\boldsymbol{\phi}}$ denotes head root orientation and its velocity, $\boldsymbol{\theta}$ denotes the FLAME face pose parameters,
and $\textbf{J}$, $\dot{\textbf{J}}$ denotes facial joint and its velocity, respectively.

The generative facial motion prior is trained to predict the facial motion state, $\textbf{x}_{t+1}$, given the current state $\textbf{x}_{t}$ as a condition. 
We consider the NeuFace holdout test split as the real motion distribution and compute the FD
for the generated motions.
We do not consider VOCASET as the real motion distribution for computing FD. It is limited in diversity and naturalness, which contradicts FD's purpose of measuring the naturalness of generated motions. Our NeuFace holdout test split is much larger and more diverse than VOCASET.

\paragraph{Fine-tuning face mesh regressor}
As our dataset's another application, we improve the accuracy of the pre-trained DECA model with our NeuFace-dataset and its 3D annotations.
Specifically, we fine-tune the pre-trained DECA parameters with 
our NeuFace-dataset and the auxiliary
3D supervisions proposed in the main paper (L795-797).
During fine-tuning, we 
use an adjusted learning rate, $1\times10^{-5}$, which is ten times smaller than training DECA from scratch.
Note that there exist the DECA-coarse model and the DECA-detail model. Unfortunately, there are known issues in reproducing DECA-detail due to the absence of VGGFace2 and the training recipe (DECA GitHub issues:~\href{https://bit.ly/3jj2psn}{bit.ly/3jj2psn}, \href{https://bit.ly/3HIiVf0}{bit.ly/3HIiVf0}). 


\section{Additional experiments}
\label{sec:supp_add_exp}
In \Sref{sec:supp_emoca}, we apply NeuFace to another face mesh regressor, EMOCA~\citep{EMOCA:CVPR:2022}, showing the flexibility of our method. 
In \Sref{supp:mica}, we compare the performance of NeuFace optimization with another competing optimization method, MICA with a tracker. 
In \Sref{sec:supp_application}, we report further results of the application experiments on the main paper.


\subsection{NeuFace optimization with EMOCA}
\label{sec:supp_emoca}
Recall that we can replace the neural parameterization of face meshes with another neural model.
Specifically, we use EMOCA~\citep{EMOCA:CVPR:2022}, 
which is built upon DECA with an additional expression encoder.
We change the neural network from DECA to EMOCA
and optimize over it with our spatio-temporal and landmark losses.
In \Tref{evaluation_results} and \Sref{sec: dataset} in the main paper, we discussed about the quantitative quality of NeuFace-E-dataset.
%
%
%


\begin{figure}[h]
    \centering
    \includegraphics[width=0.7\linewidth]{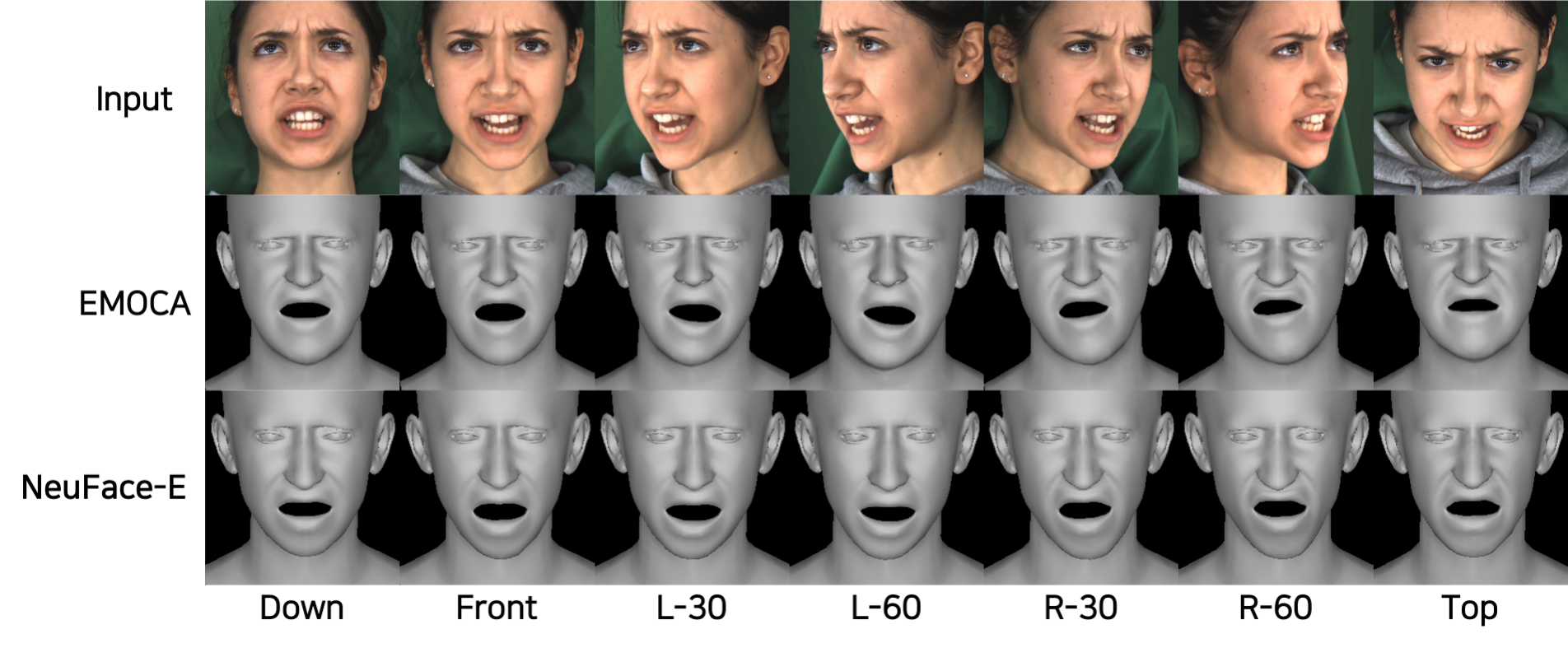}
    \caption{\textbf{Multi-view consistency: EMOCA vs. NeuFace-E.}
    We visualize the meshes obtained by EMOCA~\citep{EMOCA:CVPR:2022} and NeuFace-E.
    By aligning the meshes to face the same direction, we can clearly notice that
NeuFace-E obtains multi-view consistent meshes compared to EMOCA.
    }
    \label{fig:emoca}
\end{figure}

Qualitatively, we visualize the rendered meshes over different views as in \Fref{fig:emoca}.  NeuFace-E-dataset (3\emph{rd} row) contains multi-view consistent meshes over views compared to the meshes obtained by EMOCA inference (2\emph{nd} row). Specifically, EMOCA produces huge discrepancies in mouth area across the views, while our method is more consistent. Moreover, 
our method reconstructs more accurate meshes, especially for the shape of the nose and face contour.
%

\begin{wraptable}{R}{0.55\linewidth} 
  \centering
  \resizebox{\linewidth}{!}{
  \begin{tabular}{l cccc}
  \toprule
   & CVD $\downarrow$ & $\text{MSI}_\text{3D}$ $\uparrow$ & NME $\downarrow$ & Optim.~time~(7 views)\\
  \cmidrule(lr){1-5} 
  MICA${+}$T & 0.049 & \textbf{0.349} & 2.98 & $60~\text{min.}$\\
  \textbf{NeuFace-D-dataset} & \textbf{0.0094}  & 0.277 & \textbf{2.58} & $\mathbf{8}~\textbf{min.}$\\
  \bottomrule
  \end{tabular}}
  \caption{Evaluation on MEAD: MICA vs. NeuFace}
  \label{tab:mica}
\end{wraptable}

\subsection{NeuFace vs. MICA${+}$Tracker}\label{supp:mica}
We verify the favorable quality of NeuFace-datasets by comparing it with 
the meshes obtained by the state-of-the-art method, MICA with a tracker~\citep{MICA:ECCV2022} (MICA${+}$T).
%
%
MICA${+}$T jointly optimizes 3DMM, cameras, and textures with landmark and photometric losses, and statistic regularizers.
%
In \Tref{tab:mica}, NeuFace performs better than MICA${+}$T in CVD \& NME with comparable MSI.
Also, faster optimization makes NeuFace preferable when annotating large-scale videos.

\subsection{Facial motion prior learned from NeuFace-dataset}
\label{sec:supp_application}
In \Sref{sec:humor} of the main paper, we trained the facial motion prior model, HuMoR-Face with different training datasets: VOCASET~\citep{VOCA2019}, NeuFace\textsubscript{MEAD}, and NeuFace\textsubscript{VoxCeleb2}. 
We evaluate
HuMoR-Face models
using two metrics:
motion Fr\'echet distance (FD)~\citep{ng2022learning2listen} and average pairwise distance (APD)~\citep{mix-and-match-perturbation,rempe2021humor}.
%
FD measures the \emph{naturalness} like the FID score~\citep{fid} 
and APD measures the \emph{diversity} of generated motions.
%
%
For APD, we generate 50 long-term motions from the same initial state 
and compute the mean landmark distance between all pairs of samples.
%

\begin{table}[h!]
  \centering
  \resizebox{0.7\linewidth}{!}{
  \begin{tabular}{l c@{\quad}c@{\quad}c@{\quad}c}
    \toprule
    \textbf{HuMoR-Face trained w/} & \textbf{Scale} & \textbf{Environment} & \textbf{FD} $\downarrow$ & \textbf{APD [\emph{cm}]} $\uparrow$ \\
    \cmidrule{1-5} 
    \multicolumn{4}{l}{\textbf{Existing motion capture dataset}}\\
    \orange{VOCASET}~\citep{VOCA2019} & Small & In-the-lab & 420.92 & - \\
    \cmidrule{1-5} 
    \multicolumn{4}{l}{\textbf{Our dataset}}\\
    \greencap{NeuFace\textsubscript{MEAD}} & Large & In-the-lab & 78.99 & 3.56\\
     \blueyw{NeuFace\textsubscript{VoxCeleb2}} & Large & \underline{\textbf{In-the-wild}} & \underline{\textbf{31.32}} & \underline{\textbf{52.69}}\\
    \bottomrule
  \end{tabular}}
  \caption{{\textbf{Quantitative evaluation of learned facial motion prior.} We evaluate the \emph{naturalness} and \emph{diversity} of generated long-term motions from different motion prior models.
  HuMoR-Face trained with existing facial motion capture dataset, \eg, VOCASET~\citep{VOCA2019}, fail to generate natural and diverse facial motions.}}
  \label{tab:humor}
\end{table}

HuMoR-Face models trained with large-scale and diverse motions, \ie, the NeuFace-dataset, show superior performance in 
\emph{naturalness} and \emph{diversity} (see \Tref{tab:humor}). 
Specifically, the 
HuMoR-Face trained with NeuFace\textsubscript{VoxCeleb2} shows substantial enhancement on APD.
%
APD is not reported for 
the 
HuMoR-Face
trained with VOCASET, 
since the model fails to generate realistic motion.
%
Please check the generated motion comparisons in the supplementary video.

\section{More dataset samples}
\label{sec:supp_dataset}
\begin{figure*}[htbp]
  \centering
   \includegraphics[width=0.75\linewidth]{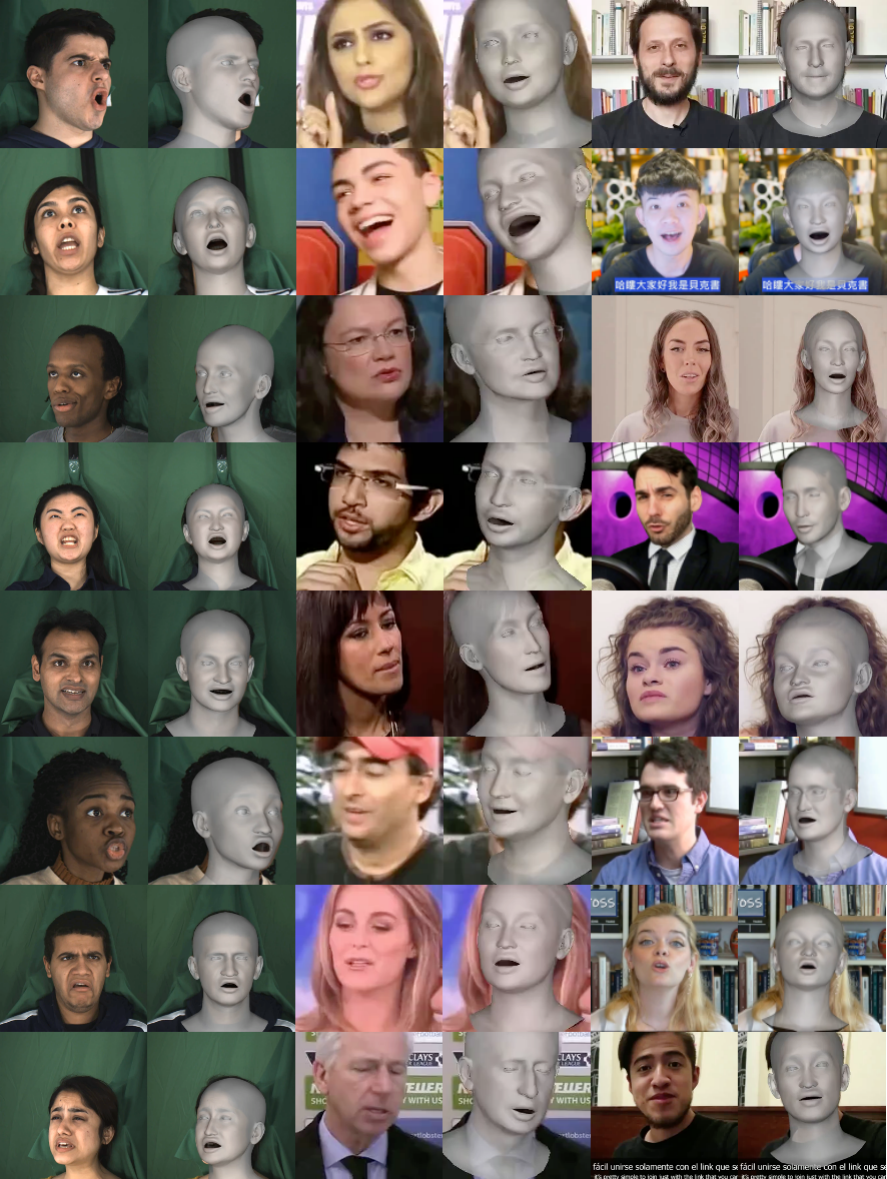}

   \caption{\textbf{NeuFace-dataset} is the large-scale 3D face video dataset containing 3DMM annotations for faces with diverse ethnicity, gender, emotions, and actions. 
   %
   Please refer to our \href{https://neuface-dataset.github.io}{project page} for the dynamic face visualizations.}
   \label{fig:supp_dataset}
\end{figure*}

%
We present more qualitative samples of 
NeuFace-dataset, with diverse identities and visual features (see \Fref{fig:supp_dataset}).
%
Since we cannot deliver expressive and temporally smooth facial motion in images, we strongly recommend seeing video visualizations for NeuFace-dataset, in the supplementary videos.

\end{document}